\documentclass{article} % For LaTeX2e
\usepackage{iclr2027_conference,times}

\usepackage{amsmath,amsfonts,bm}

\def\eqref#1{equation~\ref{#1}}
\def\1{\bm{1}}

\DeclareMathAlphabet{\mathsfit}{\encodingdefault}{\sfdefault}{m}{sl}
\SetMathAlphabet{\mathsfit}{bold}{\encodingdefault}{\sfdefault}{bx}{n}

\usepackage{hyperref}
\hypersetup{hidelinks}
\usepackage{url}            % simple URL typesetting
\usepackage{booktabs}       % professional-quality tables
\usepackage{amsfonts}       % blackboard math symbols
\usepackage{nicefrac}       % compact symbols for 1/2, etc.
\usepackage{microtype}      % microtypography
\usepackage{xcolor}         % colors
\usepackage{enumitem}
\usepackage{graphicx}
\usepackage{multirow}
\usepackage{wrapfig}
\usepackage{subcaption}
\usepackage{adjustbox}
\usepackage{amsmath}
\usepackage{amssymb}        % \varnothing and other symbols
\usepackage{float}

\title{ReVision3D: Attribution-Guided Recursive Self-Improvement for 3D Medical Perception}

\author{Ho Hin Lee \thanks{Corresponding Author: ho.hin.lee@vanderbilt.edu} \\
Vanderbilt University 
\And
Yuyin Zhou \\
University of California, Santa Cruz 
\And
Yannan Yu \\
University of California, San Francisco 
\And
Shi Gu \\
Zhejiang University 
\And
Yifan Wu \\
University of Pennsylvania
}

\iclrfinalcopy % Uncomment for camera-ready version, but NOT for submission.
\begin{document}

\maketitle

\begin{abstract}
Recursive self-improvement (RSI) offers a promising path for overcoming the limited visual capability of current medical imaging agents. Yet applying RSI to volumetric imaging remains difficult: failures can arise from acquisition, perception, training recipe, or downstream inference, while self-generated feedback and logged trajectories provide little guidance on which component should change. We introduce \textbf{ReVision3D}, an RSI system that leverages 3D volumes with spatially grounded annotations to determine where visual evidence is lost and recursively improve the corresponding visual capability. A frozen language-model designer proposes revisions to acquisition, perception, training, or inference, while the verifier and system-level objective remain fixed. Our key insight is that an annotated volume forms an \textbf{exact replay world for view rendering and spatial verification}: unvisited views can be rendered on demand, and localized predictions can be checked directly against reference masks. This grounded feedback directs targeted revision, while only changes that improve beyond measured seed noise are retained. Each accepted change triggers renewed attribution, allowing the dominant bottleneck to shift across rounds. On abdominal CT, attribution identifies perception as the dominant remaining limitation. Revising that level enables ReVision3D to achieve 79\% liver recall and 83\% kidney recall at under 0.4 false positives per patient, outperforming the evaluated frozen multimodal foundation models, with the largest gains on small lesions. Our framework is public available with interactive demo in this project page: \url{https://leeh43.github.io/ReVision3D/}
\end{abstract}

\section{Introduction}
Multimodal foundation models enable medical imaging agents to reason over images, reports, and clinical context, yet fluent reasoning does not guarantee reliable visual understanding. Reading a three-dimensional scan requires acquiring relevant evidence, recognizing subtle findings when visible, and reconciling observations across the volume. Evidence that is never rendered or perceived cannot be recovered by better downstream reasoning. This limitation is particularly important in volumetric imaging, where a model typically observes only a small fraction of the available image space. Recursive self-improvement (RSI) offers a natural path toward improving how these systems see. Recent efforts suggest a promising direction toward medical agents that
improve from diagnostic experience and progressively acquire new capabilities
~\citep{li2024agent,fan2026evolving,shen2026evo}.
Yet self-improvement of the \textbf{visual reading process itself} raises a
different question: \textbf{what part of the visual system should improve?} A missed finding may result from failed acquisition, weak perception, inconsistent detection, or downstream aggregation, and each failure mode calls for a different intervention. Improving the wrong component can therefore leave the dominant bottleneck unchanged.

\begin{figure*}[t]
    \centering
    \includegraphics[width=0.88\textwidth]{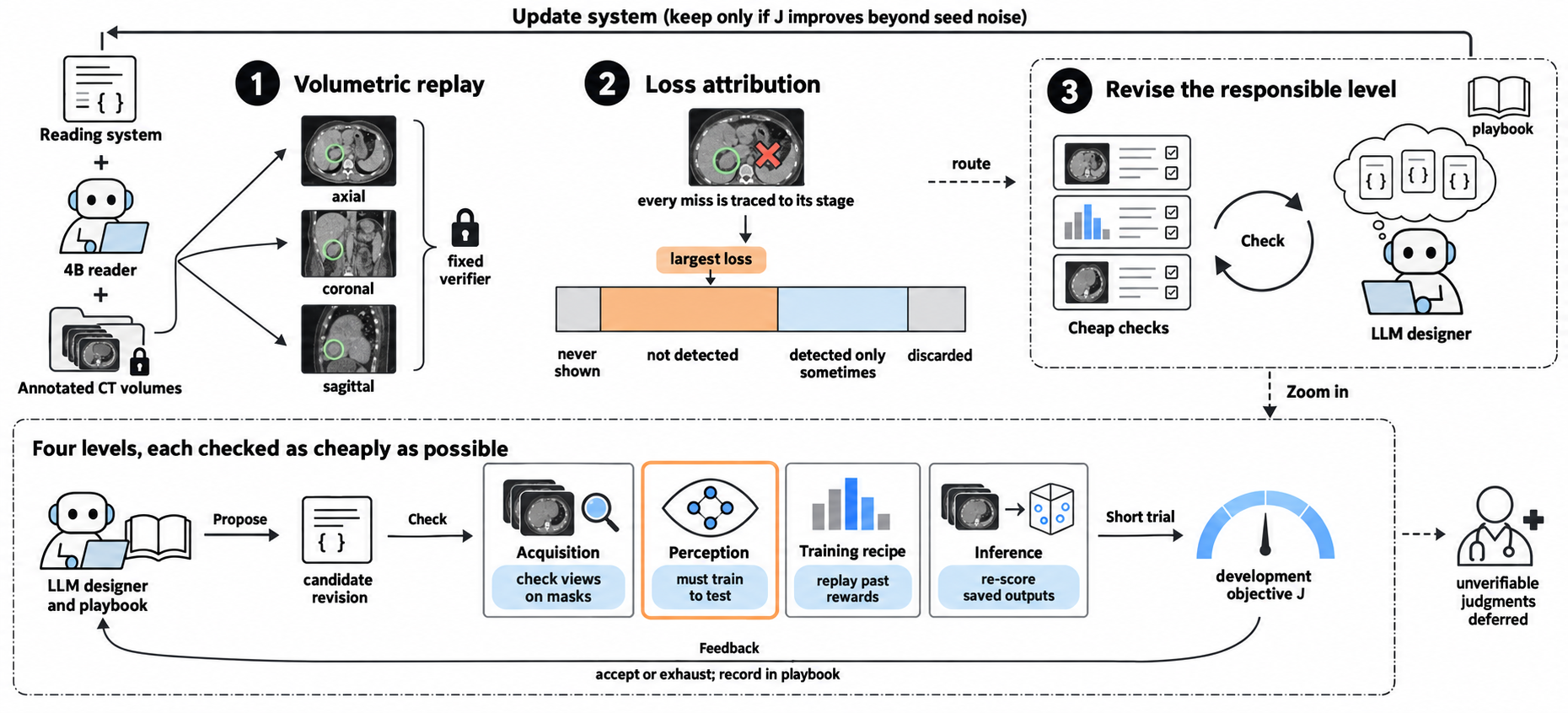}
    \caption{
    \textbf{Overview of ReVision3D.}
    \textbf{(1) Volumetric replay} renders alternative CT views and verifies localized predictions using a fixed mask-grounded verifier.
    \textbf{(2) Loss attribution} traces each missed lesion to the stage where evidence is lost.
    \textbf{(3) Revision} targets acquisition, perception, training recipe, or inference using a frozen LLM designer and replay-based checks where available.
    Changes are retained only when the fixed development objective $J$ improves beyond measured seed noise, after which attribution is recomputed.
    }
    \label{fig:method}
\end{figure*}

Conventional self-improvement provides limited feedback for answering this question. Logged trajectories reveal only what the agent inspected, so they cannot determine whether an unvisited slice or crop would have exposed a missed finding. Moreover, self-generated judgments likewise do not independently establish whether visual evidence was localized correctly. Effective visual RSI therefore requires both \textbf{attribution}, to identify where evidence is lost, and \textbf{external verification}, to determine whether a proposed revision genuinely improves visual capability. Volumetric medical imaging provides unusual structure for both. Given a scan and reference masks, previously unvisited views can be rendered on demand, their lesion coverage computed directly from geometry, and localized predictions mapped into a shared 3D coordinate system for verification. We therefore treat an annotated scan as an \textbf{exact replay world for rendering and localization}: a reusable environment for counterfactual view evaluation and grounded perception feedback without relying on the reader to judge itself.

Building on this formulation, we introduce \textbf{ReVision3D}, a framework for \textbf{attribution-guided recursive self-improvement} of medical visual capability (Figure~\ref{fig:method}). ReVision3D traces missed lesions through the visual pipeline and directs a frozen language-model designer toward one of four intervention levels: \textbf{acquisition}, \textbf{perception}, \textbf{training recipe}, or \textbf{inference}. Candidate revisions are screened through replay whenever possible, while perception changes require actual training. All candidates are then evaluated under a fixed verifier and system-level objective, and retained only when their gains exceed measured variability. Each accepted change triggers renewed attribution, allowing the dominant bottleneck to shift across rounds. We study this process on abdominal CT lesion localization, a controlled and spatially verifiable instance of medical visual perception. Recipe-level revisions improve how existing detections are used but do not move the visual ceiling; attribution instead identifies perception as the dominant remaining limitation. Revising the visual representation substantially improves recall for both liver lesions and kidney tumors, including small lesions, while inference over successive readers recovers complementary evidence unavailable to any single reader. The loop also rejects revisions that remain within seed variability rather than forcing monotonic progress. Our contributions are threefold:
\begin{itemize}
    \item \textbf{Volumetric replay.} We formulate annotated volumes as exact replay environments for rendering previously unvisited views and grounding localized predictions in a shared 3D coordinate system.
    \item \textbf{Attribution-guided RSI.} We attribute failures across acquisition, perception, training, and inference and direct improvement toward the corresponding intervention space while keeping the verifier, system-level objective, and acceptance rule fixed.
    \item \textbf{Bounded visual self-improvement.} On abdominal CT, we show that different interventions move different failure modes, that perception adaptation improves liver and kidney recall after recipe-level progress saturates, and that the loop can reject unsupported changes and expose limitations outside its current intervention space.
\end{itemize}

\section{Related Work}
\label{sec:related_work}
\textbf{Medical imaging agents and grounded perception.} Volumetric medical agents extend multimodal foundation models to CT through slice rendering and segmentation or measurement tools, often leaving the image reader itself fixed~\citep{wang20263dmedagent,mao2026ct}. Complementary work trains medical readers with scan-grounded rewards~\citep{lin2026regulating,li2026mrl,li2026medreason}, while active-perception methods learn cropping or zooming policies to acquire informative observations~\citep{zheng2026deepeyes,liu2026zoomearth}. ReVision3D connects these directions by treating acquisition, perception, training, and inference as distinct intervention spaces and using volumetric replay to evaluate counterfactual views without reader inference.

\textbf{Recursive self-improvement and self-evolving agents.}
Prior work on recursive improvement spans executable artifact search, where language models iteratively propose and evaluate programs or algorithms against external objectives~\citep{romera2024mathematical,shojaee2025llm,novikov2025alphaevolve}, and agent-level self-evolution, where accumulated experience is used to refine memory, skills, tools, or execution policies~\citep{li2024agent,fan2026evolving,shen2026evo}. Dream-RSI further uses recorded discovery trees as replay environments for improving exploration policies~\citep{zheng2026dream}. Recent medical systems extend this direction by turning diagnostic experience into reusable clinical capabilities and persistent tools~\citep{fan2026evolving}. ReVision3D differs in both the \emph{object of improvement} and the \emph{source of feedback}: it recursively revises the visual reading process itself, while an annotated 3D volume provides counterfactual observations that need not appear in any logged trajectory. Volumetric replay and loss attribution then determine whether the next revision should target acquisition, perception, training, or inference under a fixed external verifier.

\section{ReVision3D: Recursive Self-Improvement through Volumetric Replay}
\label{sec:method}
\label{sec:motivation}

\paragraph{Motivation: annotated scans as replay worlds.}
A stored medical volume contains more information than the views an agent happened to inspect. Given the acquisition geometry, one can render any admissible slice or crop after the fact and determine which annotated lesions that view exposes. This resembles model-based reinforcement learning, where simulated experience evaluates policies without repeated interaction~\citep{hafner2023mastering}, and replay-based RSI, where recorded discovery trees support counterfactual search over previously explored branches~\citep{zheng2026dream}. A volume with reference masks provides a stronger object for the spatially verifiable part of medical perception: rendering is known rather than learned, unvisited views remain available, and localized predictions can be checked against fixed geometry. It therefore supports two forms of replay needed for visual self-improvement: counterfactual replay of \textbf{where to look} and grounded replay of \textbf{what was localized}.

As shown in Figure~\ref{fig:method}, ReVision3D alternates between \textbf{attributing failures} in replay and \textbf{revising} the medical reading system. The system $\Sigma=(\psi,\varphi,\theta,s)$ contains a view-selection program $\psi$, a perception configuration $\varphi$, reader weights $\theta$, and an inference program $s$. Each revision cycle consists of three steps. Replay first attributes each missed lesion to the stage where evidence is lost. A frozen language-model designer then revises the corresponding component ($\psi$, $\varphi$, $s$, or the training recipe $\eta$ that produces $\theta$). Finally, a fixed evaluator determines whether the revision is retained. Only $\Sigma$ is allowed to change. The verifier, reference masks, system objective, and acceptance rule remain fixed, and the designer cannot inspect evaluation examples or modify what scores it.

\paragraph{Formalizing volumetric replay.}
A replay world is $w=(x,m)$, where $x$ is a CT volume and $m$ its reference lesion masks. A view $a$ (a plane and position, or a zoomed crop) is rendered deterministically as $I_a=\mathcal R(x,a)$, together with a known map $\phi_a$ from image to physical coordinates. When the reader reports a location $\mathbf p_i$ on view $a_i$, the finding is lifted to $\mathbf z_i=\phi_{a_i}(\mathbf p_i)$ in a shared 3D frame, where findings from different slices and planes are reconciled into lesion-level candidates. Deployment and replay use the same read interface, while replay additionally renders views that were never requested and evaluates localized findings against $m$. The replay world is therefore exact for rendering, geometric exposure, and spatial verification, but it does not assume the reader's response to an unvisited view without actually running the reader.

This structure provides two complementary forms of grounded feedback: \emph{counterfactual exposure}, which asks whether a view would show a lesion geometrically, and \emph{grounded verification}, which asks whether a reported location corresponds to a reference lesion. We use the same fixed spatial tolerance $\tau=5$\,mm for both.

\paragraph{Counterfactual exposure.}
For a view set $A$, let $\mathrm{Vis}(A)$ denote the lesions geometrically exposed by at least one view in $A$. A lesion is considered exposed when a view plane intersects its extent or passes within $\tau$ of it. Because $\mathrm{Vis}(A)$ depends only on scan geometry and reference masks, a view-selection program can be evaluated for the lesions it would expose without invoking the reader. Exposure and detection are therefore distinct: a lesion may be geometrically covered by a rendered view yet still be missed by the reader.

\paragraph{Grounded verification.}
For prediction grounding, each lifted location $\mathbf z_i$ is matched to the nearest reference lesion within the same tolerance $\tau$, yielding distinct hits $H(y)$, duplicates $D(y)$, and ungrounded findings $F(y)$. For a parseable output $y$, the training reward is
\begin{equation}
R_\lambda(y;w,A)=\underbrace{|H(y)|}_{\text{lesions found}}
-\underbrace{\lambda_d D(y)}_{\text{duplicates}}
-\underbrace{\lambda_f F(y)}_{\text{false findings}}
+\underbrace{b_0\,\mathbf 1\!\left[y=\varnothing\land\mathrm{Vis}(A)=\varnothing\right]}_{\text{correct silence}},
\label{eq:reward}
\end{equation}
with $b_0=0.5$ and $\lambda=(\lambda_d,\lambda_f)\in[0.1,1]^2$. The verifier is fixed and a training recipe may tune only the bounded weights $\lambda$, which shift the precision-recall trade-off but cannot redefine a correct localization. Reference masks are used only by replay and verification and are never shown to the reader.

\paragraph{System objective.}
Grounded verification defines whether a predicted location is correct, and
$R_\lambda$ uses that signal to train the reader. System promotion, however,
is determined by a separate fixed objective. Thresholding the inference score
produces operating points $(\mathrm{Rec}_k,\mathrm{FP}_k)$. For false-positive
budgets $\mathcal B=\{0.25,0.5,1,2\}$ per patient, we define
\begin{equation}
J(\Sigma)=\frac{1}{|\mathcal B|}\sum_{b\in\mathcal B}
\max_{k:\,\mathrm{FP}_k\le b}\mathrm{Rec}_k ,
\label{eq:objective}
\end{equation}
macro-averaged over organs. Unlike the training reward, $J$ has no editable
parameters: every candidate revision is judged by the same criterion.
Development patients determine promotion. Final multi-organ evaluation occurs only
after all choices are frozen. Appendix~\ref{app:setup} documents overlap with earlier component-study test cohorts.

\paragraph{Failure attribution.}
Replay lets us trace each missed lesion through the stages of the deployed
visual pipeline. For lesion $j$, let $E_j$ indicate that it is geometrically
exposed by the deployed views (or is hit by any recorded read),
$\widehat Q_j^G$ that it is detected either by the deployed read or by at least
one of $G$ additional sampled reads of the same views, $D_j$ that it is
detected by the deployed read, and $K_j^b$ that the resulting candidate is
retained at false-positive budget $b$. By construction,
$K_j^b\le D_j\le \widehat Q_j^G\le E_j$, so
\begin{equation}
1-\mathrm{Rec}_b=
\underbrace{\mathbb E[1-E_j]}_{\text{not shown}}
+\underbrace{\mathbb E[E_j-\widehat Q_j^G]}_{\text{shown, never detected}}
+\underbrace{\mathbb E[\widehat Q_j^G-D_j]}_{\text{detected only when sampled}}
+\underbrace{\mathbb E[D_j-K_j^b]}_{\text{detected, then discarded}} .
\label{eq:attribution}
\end{equation}
The decomposition itself is an exact, non-negative identity. The distinction
between the two middle terms is sample-based: with finite $G$, failure to
observe a hit cannot establish that a lesion is truly unreachable by the
reader. On development and test patients we therefore do not use additional
sampling and report these terms jointly as the \emph{never-detected} share,
$\mathbb E[E_j-D_j]$. Attribution is used to identify which intervention
space remains limiting and should not be interpreted as a causal claim.

\paragraph{Four levels of revision.}
Each term in Eq.~(\ref{eq:attribution}) maps to a distinct intervention.
\emph{Not shown $\rightarrow$ acquisition:} $\psi$ requests at most $B$ additional views, and candidate programs are screened using geometric exposure $\mathrm{Vis}$ without invoking the reader.
\emph{Shown but not detected $\rightarrow$ perception:} $\varphi$ selects the adapted vision modules, LoRA rank~\citep{hu2021lora}, and input resolution. Candidates require actual training with a supervised warm start followed by Eq.~(\ref{eq:reward}), while pretrained weights remain frozen.
\emph{Detected inconsistently $\rightarrow$ training recipe:} group-relative optimization~\citep{shao2024deepseekmath} learns only from prompts with non-identical sampled rewards. A \textbf{learnability census} $\mathcal C_\theta$ stores these rewards, and for recipe distribution $q_\eta$ replay estimates
\begin{equation}
\widehat S(\eta)=\sum_u q_\eta(u)
\left[1-\sum_v \widehat p_u(v)^G\right],
\label{eq:census}
\end{equation}
the expected fraction of non-degenerate groups, used only as a screening proxy.
\emph{Discarded $\rightarrow$ inference:} $s$ scores merged 3D candidates from label-free features such as cross-slice support and per-reader contributions. Cached outputs allow candidate programs to be evaluated directly on $J$ without retraining.

\paragraph{Recursive improvement.}
At round $t$, attribution measures the incumbent's not-shown, never-detected,
and discarded losses. The never-detected term contains two qualitatively
different cases: (i) lesions that the current reader can detect on some sampled
reads, which may still benefit from training, and (ii) lesions that are never
detected, which more strongly suggests a perception bottleneck. We estimate
this split using the learnability census. Let
$r_t=n_{\mathrm{front}}/(n_{\mathrm{front}}+n_{\mathrm{never}})$ denote the
fraction of lesion-bearing prompts that lie on the learning frontier among
those not always detected. We then define
\begin{equation}
\widehat L_{\mathrm{perc}}=(1-r_t)L_{\mathrm{undet}},\qquad
\widehat L_{\mathrm{pol}}=r_tL_{\mathrm{undet}},\qquad
\ell_t=\arg\max_{\ell\notin\mathcal X_t}\widehat L_\ell ,
\label{eq:routing}
\end{equation}
where $\mathcal X_t$ contains exhausted intervention levels.
Thus, a larger learning-frontier share assigns more of the undetected loss to
training, whereas a larger never-hit share directs the search toward
perception. This split is a routing heuristic rather than an exact
decomposition.

The designer then proposes revisions only within level $\ell_t$, using replay
screens where available. Screens rank candidates but never promote them.
Training revisions are evaluated on matched development patients and seeds
$\xi$ and accepted only if $\overline{\Delta J}>\nu$ and $\min_\xi\Delta J^{(\xi)}>0$, where $\nu$ is the incumbent's seed range. Inference revisions require
$P(\Delta J>0)\ge0.9$ under a paired bootstrap. An accepted revision updates
$\Sigma$ and triggers new attribution, otherwise the level is eventually
marked exhausted. Across rounds, an evidence-backed \textbf{playbook} carries
successful and failed hypotheses forward. The routing rule in
Eq.~(\ref{eq:routing}) was formulated retrospectively and did not schedule the
experiments reported here.

% \paragraph{Deployment and scope.}
% Reference masks are used only for training, replay, and evaluation and are unavailable at deployment. Budgeted reads let $\psi$ request additional views after an initial survey, whereas dense reads process axial slices at fixed spacing and reconcile findings in 3D. ReVision3D optimizes only spatially verifiable localization; unsupported clinical judgments remain outside the objective and are deferred to clinicians, while the downstream reasoning agent is unchanged.

\section{Experimental Design}
\label{sec:experiments}

We evaluate ReVision3D on abdominal CT lesion localization using LiTS~\citep{bilic2023liver} for liver lesions and KiTS23~\citep{heller2023kits21,myronenko2023automated} for kidney tumors. A prediction counts as a hit if its localized point lies within 5~mm of a reference lesion mask and the primary analysis includes lesions of at least 1~mL. The reader is Qwen3-VL-4B-Instruct~\citep{bai2025qwen3}, and the default designer is Claude Opus 5. All model, program, and threshold selections are performed with development patients. Final multi-organ evaluation occurs after all choices are frozen. Appendix~\ref{app:setup} documents overlap with earlier component-study test cohorts and gives full splits, rendering details, and optimization settings.

Our primary baselines are thirteen frozen frontier and open vision--language models. Each receives the same rendered images, captions, prompt, and output grammar as our reader on a uniform survey of eight axial slices. Eight slices match the observation budget of our budgeted acquisition study and geometrically expose all liver lesions $\geq1$~mL under the 5~mm tolerance, even lesion-guided oracle slices raise no frontier reader above 0.43 recall (Appendix~\ref{app:frozen}). ReVision3D uses a \emph{dense read}: axial slices at 5~mm spacing are processed one per prompt and neighboring findings are merged into 3D candidates. A candidate's \emph{support} is the number of slices reporting it, so varying the support threshold traces a recall--false-positive operating curve. To separate protocol from perception, we also evaluate the strongest matched frontier baseline, Gemini 3.8 Flash, under the same dense read and compare our reader variants under an identical protocol.

For the survey baselines we report recall at $\leq0.5$ false positives per patient, overall and by lesion volume (1--5, 5--20, 20--100, and $>100$~mL). System revisions are promoted using $J$ from Eq.~(\ref{eq:objective}). Confidence intervals use patient-level bootstrap resampling, and paired systems are compared with paired bootstraps.

\begin{table}[t]
\centering
\footnotesize
\setlength{\tabcolsep}{3.4pt}
\caption{\textbf{Recall by lesion volume on the liver and kidney test patients.} Recall at $\leq$0.5 false positives per patient. Frontier models read 8 uniform slices, ours every slice (same-protocol comparison: Appendix Table~\ref{tab:sizes_protocol}). Bold: best in column; underlined: best frontier model.}
\label{tab:sizes}
\begin{adjustbox}{max width=\linewidth}
\begin{tabular}{l ccccc ccccc}
\toprule
 & \multicolumn{5}{c}{Liver: 31 patients, 111 lesions} & \multicolumn{5}{c}{Kidney: 70 patients, 81 lesions} \\
\cmidrule(lr){2-6}\cmidrule(lr){7-11}
 & \multicolumn{4}{c}{Lesion volume (mL)} & & \multicolumn{4}{c}{Lesion volume (mL)} & \\
\cmidrule(lr){2-5}\cmidrule(lr){7-10}
Reader & 1--5 & 5--20 & 20--100 & $>$100 & All & 1--5 & 5--20 & 20--100 & $>$100 & All \\
{\scriptsize\textit{Number of lesions}} & {\scriptsize52} & {\scriptsize31} & {\scriptsize17} & {\scriptsize11} & {\scriptsize111} & {\scriptsize14} & {\scriptsize24} & {\scriptsize25} & {\scriptsize18} & {\scriptsize81} \\
\midrule
Mistral Large 3 & 0.06 & 0.00 & 0.06 & 0.82 & 0.12 & 0.00 & 0.04 & 0.00 & 0.61 & 0.15 \\
Qwen3.5-397B & 0.00 & 0.10 & 0.06 & 0.36 & 0.07 & 0.00 & 0.08 & 0.56 & 0.94 & 0.41 \\
Gemini 3.1 Pro & 0.12 & 0.42 & \underline{0.71} & 0.73 & 0.35 & \underline{0.21} & 0.17 & 0.68 & \textbf{1.00} & 0.52 \\
Gemma 4 31B & 0.02 & 0.10 & 0.41 & 0.82 & 0.18 & 0.14 & 0.29 & 0.72 & 0.94 & 0.54 \\
GPT-5.5 & 0.04 & 0.13 & 0.24 & 0.64 & 0.15 & 0.00 & 0.17 & 0.60 & 0.83 & 0.42 \\
Claude Opus 5 & 0.00 & 0.10 & 0.18 & 0.45 & 0.10 & 0.00 & 0.00 & 0.28 & 0.78 & 0.26 \\
DeepSeek V4.1 Flash & 0.00 & 0.00 & 0.00 & 0.55 & 0.05 & 0.07 & 0.04 & 0.24 & 0.67 & 0.25 \\
Qwen3.8-27B & 0.00 & 0.03 & 0.24 & 0.55 & 0.10 & 0.00 & 0.04 & 0.36 & 0.83 & 0.31 \\
Qwen3.8-Max & \underline{0.13} & 0.35 & 0.47 & \textbf{0.91} & 0.32 & 0.14 & 0.17 & 0.52 & \textbf{1.00} & 0.46 \\
GLM-5.3 Flash & 0.04 & 0.10 & 0.29 & 0.64 & 0.15 & 0.07 & 0.21 & 0.32 & 0.89 & 0.37 \\
Qwen3.8-Flash & 0.02 & 0.10 & 0.18 & 0.45 & 0.11 & 0.00 & 0.21 & 0.48 & 0.94 & 0.42 \\
Gemini 3.8 Flash & 0.08 & \underline{0.48} & 0.65 & \textbf{0.91} & \underline{0.36} & \underline{0.21} & 0.46 & 0.68 & \textbf{1.00} & 0.60 \\
GPT-6 Astra & \underline{0.13} & 0.29 & 0.65 & 0.82 & 0.32 & 0.14 & \underline{0.58} & \underline{0.92} & \textbf{1.00} & \underline{0.70} \\
\midrule
\multicolumn{11}{l}{\textit{Ours: the same 4B reader, dense read (every slice); only the perception configuration differs}} \\
Frozen vision encoder & 0.06 & 0.13 & 0.41 & 0.64 & 0.19 & 0.14 & 0.29 & 0.88 & \textbf{1.00} & 0.60 \\
Adapters, 4 blocks (human-chosen) & 0.12 & 0.19 & 0.53 & 0.64 & 0.25 & 0.29 & 0.50 & 0.92 & \textbf{1.00} & 0.70 \\
\textit{ReVision3D} (designer's adapters) & \textbf{0.69} & \textbf{0.84} & \textbf{1.00} & 0.82 & \textbf{0.79} & \textbf{0.57} & \textbf{0.71} & \textbf{0.96} & \textbf{1.00} & \textbf{0.83} \\
\bottomrule
\end{tabular}
\end{adjustbox}
\end{table}

\section{Results}
\label{sec:results}

\subsection{Comparison with Frontier Models}
\label{sec:res-frontier}
\begin{figure}[t]
\centering
\includegraphics[width=0.9\linewidth]{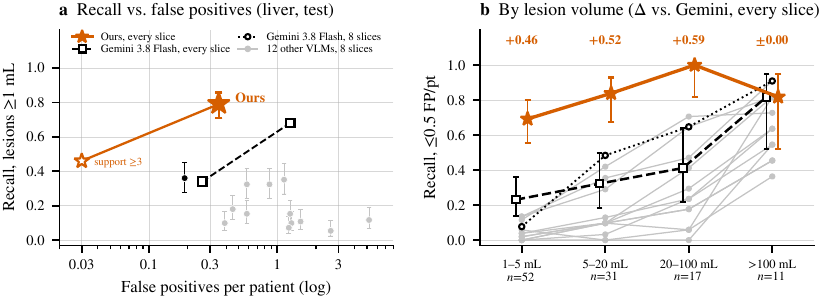}
\caption{\textbf{ReVision3D against frozen frontier models}, 31 liver test patients (111 lesions $\geq$1\,mL).
(a) Recall against false positives per patient; dense readers at support (slices reporting a finding) $\geq$3 (left point) and
$\geq$2 (right point).
(b) Recall by lesion volume at $\leq$0.5 false positives per patient; numbers above the bins are differences in recall from Gemini
3.8 Flash reading every slice under our protocol. Wilson 95\% intervals over lesions.}
\label{fig:headline}
\end{figure}
\paragraph{Main results.}
Table~\ref{tab:sizes} and Figure~\ref{fig:headline} summarize held-out
performance. At $\leq0.5$ false positives (FP) per patient, ReVision3D detects
0.79 of liver lesions (88/111), compared with 0.05--0.36 across thirteen
frozen frontier readers. Gemini 3.8 Flash is the strongest frozen reader on
liver, reaching 0.36 recall. Kidney tumors are generally easier for the
evaluated models, with frozen-reader recall ranging from 0.15 to 0.70.
ReVision3D reaches 0.83, compared with 0.70 for the strongest frozen baseline,
GPT-6 Astra. The largest gap appears for small lesions. For 1--5\,mL lesions,
ReVision3D reaches 0.69 recall on liver and 0.57 on kidney, whereas no frozen
frontier reader exceeds 0.13 and 0.21, respectively. The advantage narrows for
large lesions, where several frozen readers already approach ceiling
performance.

\paragraph{Same reading protocol.}
Reading every slice does not close the gap by itself. Under the same dense
protocol, Gemini 3.8 Flash detects 0.34 of liver lesions at $\leq0.5$ FP per
patient and reaches 0.68 recall only at 1.26 FP per patient
(Figure~\ref{fig:headline}a). Its system objective is $J=0.34$, compared with
0.71 for ReVision3D (paired difference $+0.37$ [$+0.15$, $+0.47$]). On kidney,
$J$ increases from 0.53 to 0.80 ($+0.27$ [$+0.17$, $+0.37$]). Thus, additional
slices alone do not explain the improvement: at a comparable false-positive
rate, the frozen reader performs similarly under dense and survey reading
(0.34 versus 0.36 recall on liver, Appendix Table~\ref{tab:sizes_protocol}).

\paragraph{Where the gain comes from.}
The last three rows of Table~\ref{tab:sizes} hold the reader family, training
data, and dense protocol fixed while changing only the perception
configuration. With the vision encoder frozen, our reader detects 0.19 of
liver lesions, adapting four vision blocks raises recall to 0.25, while the
designer-selected configuration reaches 0.79. For 1-5\,mL lesions, recall
rises from 0.06 to 0.12 to 0.69 on liver and from 0.14 to 0.29 to 0.57 on
kidney. This controlled comparison isolates perception as the largest source
of improvement, consistent with the attribution analysis in
Section~\ref{sec:res-dynamics}.

\subsection{Recursive Improvement Trajectory}
\label{sec:res-dynamics}

\begin{figure}[t]
\centering
\includegraphics[width=0.88\linewidth]{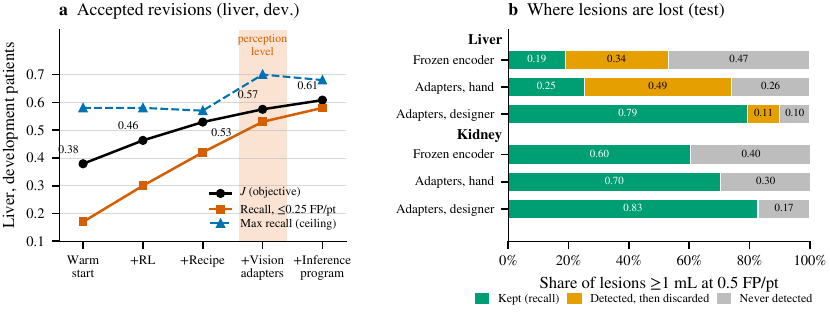}
\caption{\textbf{Attribution shows where improvement stops and which level moves it.}
(a) The liver system after each accepted revision (20 development patients, dense read): objective $J$, recall at $\leq$0.25 false
positives per patient, and the maximum reachable recall. (b) Where lesions $\geq$1\,mL are lost at 0.5 false positives per patient
(Eq.~(\ref{eq:attribution})) for the three perception configurations in the single test evaluation }
\label{fig:trajectory_attribution}
\end{figure}

\paragraph{Improvement trajectory.}
Figure~\ref{fig:trajectory_attribution}a follows the liver system across
successive accepted revisions. Reinforcement learning with the mask-grounded
reward raises $J$ from 0.379 to 0.463, and the learnability-aware recipe further
raises it to 0.529. These gains primarily improve precision: recall at
0.25 false positives per patient increases from 0.17 to 0.42, while the maximum
reachable recall remains nearly unchanged at 0.57--0.58. Recipe-level progress
then saturates. Across two recipe searches, none of fourteen designer proposals
exceeds its incumbent by more than the measured noise floor after full training
(Appendix~\ref{app:recipe}).
 
\paragraph{The bottleneck shifts to perception.}
Recipe-level improvement leaves a large never-detected term. With the frozen
encoder, the share is 0.42/0.45 on liver/kidney development and 0.47/0.40 on
test (Figure~\ref{fig:trajectory_attribution}b). Prompts whose sampled outputs
never hit these lesions provide no group-relative learning signal. Applied
retrospectively, the routing rule in Eq.~(\ref{eq:routing}) therefore assigns
the dominant remaining loss to perception. Adapting the vision encoder raises
the reachable recall ceiling from 0.57 to 0.70 and reduces the never-detected
share to 0.25/0.14 on development and 0.10/0.17 on test. The subsequent
inference revision combines complementary evidence from successive readers,
reaching 0.58 recall at 0.25 false positives per patient and $J=0.608$.

\subsection{Perception Level and Test Evaluation}
\label{sec:res-test}

\begin{figure}[t]
\centering
\includegraphics[width=\linewidth]{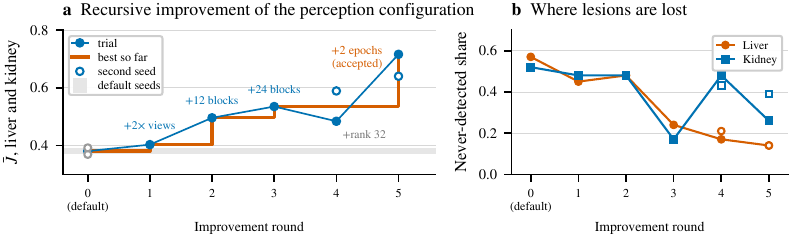}
\caption{
\textbf{Perception-level self-improvement.}
\textbf{(a)} Recursive search over perception configurations, where $\bar J$ is the mean liver and kidney development objective. The orange line tracks the best configuration found so far and hollow markers show second-seed repeats.
\textbf{(b)} The corresponding never-detected lesion share for liver and kidney.
Adapting more vision blocks substantially improves $\bar J$ and reduces the never-detected share, while the two-epoch configuration is ultimately accepted.
}
\label{fig:perception_round_fig}
\end{figure}

\begin{table}[t]
\centering
\footnotesize
\setlength{\tabcolsep}{3.6pt}
\caption{\textbf{Perception and inference revisions on liver and kidney.} Paired entries denote seeds $s_0$\,/\,$s_1$; single entries report the available evaluation. Bottom rows compare a fixed seed program that merges candidates from successive accepted readers with the inference program selected on development ($P(\Delta J{>}0)=0.96$).}
\label{tab:multi_organ_main}
\begin{adjustbox}{max width=\linewidth}
\begin{tabular}{ll ccc ccc}
\toprule
 & & \multicolumn{3}{c}{Development $J$ (selection)} & \multicolumn{3}{c}{Test $J$ (once)} \\
\cmidrule(lr){3-5}\cmidrule(lr){6-8}
Perception configuration & Chosen by & Liver & Kidney & Mean & Liver & Kidney & Mean \\
\midrule
Frozen encoder {\scriptsize(language adapters only)} & fixed & 0.48 & 0.53 & 0.51 & 0.28 & 0.57 & 0.43 \\
Adapters, hand {\scriptsize(4 blocks, rank 16, 1$\times$)} & hand & 0.50 / 0.47 & 0.67 / 0.76 & 0.58 / 0.61 & 0.39 & 0.68 & 0.54 \\
Adapters, designer {\scriptsize(24 blocks, rank 32, 2$\times$)} & designer & 0.71 / 0.70 & 0.86 / 0.80 & 0.79 / 0.75 & 0.71 / 0.54 & 0.80 / 0.78 & 0.75 / 0.66 \\
\midrule
\multicolumn{8}{l}{\textit{Inference level on the designer's configuration (program over successive accepted readers)}} \\
+ successive readers merged {\scriptsize(seed program)} & fixed & 0.68 & 0.80 & 0.74 & 0.78 & 0.80 & 0.79 \\
+ designer's inference program {\scriptsize(Claude Opus 5)} & designer & \textbf{0.75} & \textbf{0.88} & \textbf{0.82} & \textbf{0.82} & \textbf{0.83} & \textbf{0.83} \\
\bottomrule
\end{tabular}
\end{adjustbox}
\end{table}

\paragraph{Perception search.}
After recipe-level progress saturates, attribution indicates that the dominant
remaining loss lies in perception. The designer therefore searches the
perception configuration (Figure~\ref{fig:perception_round_fig}, Appendix
Table~\ref{tab:perception_round}). Doubling the input resolution produces only
a small gain, whereas adapting progressively more of the vision encoder raises
the mean liver-kidney objective $\bar J$ from 0.381 to 0.496 with 12 blocks
and 0.535 with all 24 blocks. Increasing adapter rank adds little, while
extending the stronger configuration to two training epochs reaches
$\bar J=0.716$. An independent repeat reaches 0.640, and the revision is
accepted.

Figure~\ref{fig:perception_round_fig}b shows that the same revisions also move
the attribution term that motivated the search. Broader vision adaptation
substantially reduces the never-detected share on both liver and kidney,
whereas resolution changes and shallower adaptation have smaller effects.
The accepted two-epoch configuration reduces this share to 0.14 on liver and
0.26 on kidney in its first trial. Thus, perception adaptation changes not
only the operating point but also which lesions the reader is able to detect.

\paragraph{Held-out evaluation.}
We retrain the three perception configurations on the full training set and
evaluate them once on held-out patients (Table~\ref{tab:multi_organ_main}).
The development ordering persists on test: the designer-selected configuration
achieves mean $J=0.75$ and $0.66$ across two seeds, compared with 0.54 for the
human-selected adapters and 0.43 with the vision encoder frozen. For seed
$s_0$, this corresponds to gains of $+0.21$ [$+0.14$, $+0.28$] over the
human-selected configuration and $+0.33$ [$+0.26$, $+0.41$] over the frozen
encoder.

The revised reader also provides complementary evidence for the subsequent
inference level. Merging candidates from successive readers yields test
$J=0.78$ on liver and $0.80$ on kidney. The inference program selected on
development ($P(\Delta J>0)=0.96$) reaches 0.82 and 0.83 on test. Repeating
inference-level search with four designers
from three model families yields an accepted program in all twelve runs
(Appendix Table~\ref{tab:designers}), indicating that the improvement procedure
is not tied to a single designer model.

\subsection{Ablation Analysis}
\label{sec:res-ablation}

Appendix Table~\ref{tab:ablation_main} isolates the effects of training signal,
reading protocol, training recipe, and perception configuration on the same 31
held-out liver patients.

\noindent\textbf{Training signal.}
A supervised warm start raises survey recall from 0.05 to 0.25, mainly for
lesions above 20\,mL. Reinforcement learning adds little under the same sparse
survey (0.27 recall) and does not improve 1--5\,mL lesions, while ten supervised
epochs reach 0.29 recall at 1.71 false positives per patient. RL is more useful
under dense reading, where development $J$ rises from 0.38 to 0.46, but stronger
training alone does not overcome the small-lesion bottleneck.

\noindent\textbf{Reading protocol.}
Reading every slice raises overall recall from 0.27 to 0.45 and 1--5\,mL recall
from 0.08 to 0.29, while approximately halving false positives. Confirming
detections with a second reader further raises overall recall to 0.56. Increased
coverage therefore recovers additional findings, but most small lesions remain
undetected.

\noindent\textbf{Training recipe.}
Recipe changes mainly reshape the precision--recall trade-off rather than expand
what the reader can detect. Overall recall remains within 0.41--0.49 and
1--5\,mL recall within 0.23--0.31. Training on learnable prompts from fresh
views reaches 0.49 recall versus 0.41 for the equal-compute control, while the
learning frontier contracts from 25\% to 7\%, explaining why further recipe
search saturates.

\noindent\textbf{Perception configuration.}
Changing perception produces the largest gain, especially for small lesions.
Under the same dense protocol, 1--5\,mL recall rises from 0.06 with a frozen
encoder and 0.12 with four hand-selected blocks to 0.69 with the
designer-selected configuration. Overall recall rises from 0.19 and 0.25 to
0.79. Unlike recipe revisions, perception adaptation expands the set of lesions
that the reader detects at all, consistent with the attribution analysis in
Section~\ref{sec:res-dynamics}.

\section{Discussion}
\label{sec:discussion}

ReVision3D improves medical visual capability using feedback external to the reader's own judgment. The final reader reaches 0.79 liver-lesion recall and 0.83 kidney-tumor recall at under 0.4 false positives per patient, with the largest gains on 1--5\,mL lesions. More importantly, the interventions affect different failure modes: recipe-level revisions refine existing detections, whereas perception adaptation reduces the never-detected share and raises the reachable recall ceiling. Similar inference-level outcomes across four tested designers suggest that fixed external evaluation reduces dependence on the model proposing revisions.

More broadly, recursive improvement benefits from separating \textbf{where the system fails} from \textbf{how it should be changed}. The tested interventions are not interchangeable: dense reading exposes more findings, recipe changes exploit stochastic detections, and perception adaptation raises the detection ceiling. Re-attributing failure after each accepted revision provides a mechanism for shifting search when the bottleneck moves.

\textbf{Limitations.}
The routing rule uses training-patient learnability to estimate which intervention may reduce development loss and was formulated retrospectively, so prospective routing remains to be tested on a new organ or modality. The current loop also fixes rendering choices such as windowing and neighboring-slice context, extending the intervention space to these variables is a natural next step. Additional limitations are discussed in Appendix~\ref{app:limitations}.

\section{Conclusion}
\label{sec:conclusion}

We introduced ReVision3D, an attribution-guided recursive self-improvement framework for 3D medical perception. Annotated CT volumes serve as replay environments in which unvisited views can be rendered, localized findings verified against fixed geometry, and missed lesions attributed to where evidence is lost. This enables an evidence-gated improvement loop over acquisition, perception, training, and inference. On liver and kidney CT, perception emerges as the dominant remaining limitation and produces the largest gain when revised, particularly for small lesions, while unsupported revisions are rejected. These results highlight failure attribution and externally grounded verification as useful principles for recursive improvement of medical imaging systems.

\section*{AI use statement}

Large language models are part of the method: a frozen designer (Claude Opus 5, GPT-6 Astra, Gemini 3.1 Pro, and Gemini 3.8 Flash in Appendix~\ref{app:designers}) proposes revisions of the reading system, and thirteen frontier models serve as baselines. Their roles and interfaces are described in Sections~\ref{sec:method} and~\ref{sec:experiments}. Generative AI tools were also used to assist with language editing and with scripts used to produce tables and figures from logged results. All generated material was reviewed and verified by the authors against the underlying experiments. The authors take responsibility for the full content of the paper.

% \section*{AI use statement}

% (This section is \textbf{required} and does not count toward the page limit.)

% In this work, we used generative AI tools for [tasks with required disclosure].
% We have not used generative AI tools for [other tasks with required disclosure],
% and [the rest of the required disclosure tasks] are not applicable to this work.
% Additionally, we used generative AI tools for [tasks with recommended
% disclosure]. We have reviewed all AI-assisted work. [Elaborate. For example, “we
% checked LLM-generated research ideas for potential plagiarism through a manual
% literature survey”, “LLM-generated code was verified and tested for correctness
% by 2 authors”, etc.]. We take responsibility for the final content of this work,
% including text, claims or artifacts produced with the aid of generative AI.

% See the ICLR 2027 AI Policy for Authors for more details. This statement should
% not be more than 1 page.

\subsection*{Ethics statement}

This work uses public, de-identified CT datasets (LiTS and KiTS23) under their licences, no new patient data were collected and
no human-subject experiments were run. ReVision3D is a research system for lesion localization and is not a clinical device. Reference
masks are used only for training, replay and evaluation, the system makes no claim about diagnosis, and clinical judgments that a scan
cannot verify are deferred to clinicians (Section~\ref{sec:method}). Performance on these benchmarks may not transfer to other scanners,
protocols or patient populations.

\subsection*{Reproducibility statement}

Section~\ref{sec:method} defines the reward, objective, attribution identity, screening procedures, and acceptance rule. Appendix~\ref{app:setup} gives data splits, rendering, adapter and optimization settings, the frontier-model protocol, and the self-improvement schedule. The appendix tables additionally report both accepted and rejected trials. Code, designer prompts, playbooks, and trial registries are prepared for release with the supplementary material.



\bibliography{iclr2027_conference}
\bibliographystyle{iclr2027_conference}

\newpage
\appendix
\raggedbottom
\setlength{\intextsep}{4pt plus 1pt minus 1pt}
\setlength{\textfloatsep}{6pt plus 1pt minus 2pt}
\setlength{\floatsep}{6pt plus 1pt minus 2pt}
\setlength{\abovecaptionskip}{3pt}
\setlength{\belowcaptionskip}{2pt}

\section{Terminology Used in Tables}
\label{app:terms}

For clarity, we collect the recurring terms used throughout the main manuscript and appendix tables below.

\paragraph{Reading protocols and inference.}
\begin{description}[leftmargin=1.2em,itemsep=1pt,parsep=0pt,font=\normalfont\itshape]

\item[8-slice survey.]
Eight evenly spaced axial slices presented together in a single prompt.
This is the primary protocol used to compare frozen frontier models.

\item[Dense read.]
Every axial slice is read at 5\,mm spacing, one slice per prompt.
Findings on neighbouring slices are merged into 3D candidates.
Unless otherwise stated, our trained readers are evaluated with this protocol.

\item[Lesion-guided slices.]
Slices selected to intersect every annotated lesion.
This provides an oracle estimate of the benefit available from improved view selection.

\item[Acquisition program.]
A 4-slice survey followed by up to four additional views selected by the learned
view-selection program $\psi$.

\item[Support.]
The number of rendered slices on which a 3D candidate is reported.
A threshold ``support $\geq k$'' retains only candidates observed on at least
$k$ slices; increasing $k$ typically reduces both false positives and recall.

\item[Cascade.]
An inference rule in which one reader proposes candidate findings and a second
reader must confirm each candidate within 12\,mm in the shared 3D coordinate system.

\end{description}

\paragraph{Evaluation metrics and operating points.}
\begin{description}[leftmargin=1.2em,itemsep=1pt,parsep=0pt,font=\normalfont\itshape]

\item[Operating point.]
A recall--false-positive pair obtained by fixing an inference threshold.
For example, ``recall at $\leq 0.5$ FP/patient'' denotes the highest recall
achievable while keeping false positives at or below 0.5 per patient.

\item[$R$, FP, and $J$.]
$R$ denotes lesion-level recall for lesions $\geq1$\,mL;
FP denotes false positives per patient;
$J$ is the system-level objective defined in Eq.~(\ref{eq:objective}).

\end{description}

\paragraph{Failure attribution.}
\begin{description}[leftmargin=1.2em,itemsep=1pt,parsep=0pt,font=\normalfont\itshape]

\item[Never detected.]
No evaluated read localizes the lesion.

\item[Discarded.]
The lesion is detected by at least one read but removed during downstream
aggregation or thresholding.

\item[Kept.]
The lesion is detected and retained in the final prediction.
Together, these terms correspond to the decomposition in
Eq.~(\ref{eq:attribution}).

\end{description}

\paragraph{Reader training and learning signal.}
\begin{description}[leftmargin=1.2em,itemsep=1pt,parsep=0pt,font=\normalfont\itshape]

\item[Warm start.]
Supervised initialization using mask-derived reports.

\item[RL reader.]
The reader obtained after one round of reinforcement learning with the
mask-grounded reward in Eq.~(\ref{eq:reward}).

\item[Learning frontier / learnable prompt.]
A training prompt whose sampled rewards are non-identical and therefore provide
non-zero group-relative learning signal.
This use of ``frontier'' is unrelated to \emph{frontier models}.

\item[Cached views.]
Previously rendered training views reused when constructing a learning-frontier
training pool.

\item[Fresh views.]
Newly rendered views used to recompute the learning frontier under the current reader.

\item[Learnable prompts, cached/fresh.]
Training recipes restricted to learnable prompts, using either cached or freshly
rendered views, respectively.

\end{description}

\paragraph{Recursive improvement protocol.}
\begin{description}[leftmargin=1.2em,itemsep=1pt,parsep=0pt,font=\normalfont\itshape]

\item[Seed recipe.]
The incumbent training recipe at the beginning of a recipe-level search round.

\item[Seed program.]
The incumbent inference program against which inference-level proposals are evaluated.

\item[Trial scale.]
Short training or evaluation runs on a fixed development subset used to screen
candidate revisions before full evaluation.

\item[Noise floor.]
The range $\nu$ of $J$ across repeated seeds of the
incumbent. A training revision is promoted only if its mean gain exceeds
$\nu$ and every matched seed improves.

\end{description}

\section{Experimental Details}
\label{app:setup}
\paragraph{Data and splits.}
LiTS provides 131 liver CT volumes and KiTS23 292 kidney CT volumes with voxel-level lesion masks. Lesions are the 26-connected components of
the reference masks. The \emph{liver study} (Section~\ref{sec:res-dynamics}) uses 80/20/31 patients for training, development and held-out
evaluation. The \emph{multi-organ study} (Section~\ref{sec:res-test}) uses 262/60/101 patients for training, development and test (liver
80/20/31, kidney 182/40/70).
Each study's splits are patient-disjoint. The 31 held-out liver patients form the liver test set of the
multi-organ study. They were also used for the liver-study comparisons in Appendix~\ref{app:frozen}, which informed no multi-organ decision,
and 40 of the 70 kidney test patients were used by no earlier experiment.
A fixed 30-patient subset of the development cohort (10 liver, 20 kidney) scores short designer trials, while the full 60-patient
development set governs promotion.

\paragraph{Readers and rendering.}
The reader is Qwen3-VL-4B-Instruct, a 24-block vision transformer feeding a 36-layer language model, with pretrained weights frozen.
Volumes are reoriented to RAS and cropped to an organ box expanded by 15~mm; at deployment the box is supplied by an external organ segmenter, not the reference lesion mask. Axial slices are rendered at 384 pixels on the long side in a
$[-160,240]$~HU window and captioned with plane and position. Low-rank adapters (rank 16) are trained on the language model's attention and
MLP projections. A perception configuration adds adapters to the visual mergers and a chosen number of vision blocks and may enlarge views
before the vision tower. The warm start fits two epochs of mask-derived reports (learning rate $2\times10^{-4}$, effective batch 4), while
reinforcement learning uses group-relative optimization with the Dr.~GRPO loss~\citep{liu2025understanding} (group size 8--16, temperature 0.9, learning rate
$10^{-5}$, 100--400 steps).

\paragraph{Reading protocols.}
The \emph{uniform survey} shows four or eight evenly spaced axial slices in one prompt. The \emph{budgeted read} starts from a four-slice
survey and adds at most four views requested by the acquisition program. The \emph{dense read} shows every axial slice at 5~mm, one per
prompt (about 43 per patient), and merges findings within 12~mm in 3D. A candidate's support is the number of slices that report it.
Lesion-guided slices, which pass through the annotated lesions, serve as an oracle on where to look. Beyond the reasons given in
Section~\ref{sec:experiments} for the eight-slice survey, the four-slice survey exposes only 83\% of liver lesions, and a slice within
5~mm of a small lesion may show only its edge, which is why we also report lesion-guided slices and the dense-protocol comparison.

\paragraph{Frontier readers.}
The thirteen frozen readers, in order of release, are Mistral Large 3, Qwen3.5-397B, Gemini 3.1 Pro, Gemma 4 31B, GPT-5.5, Claude Opus 5, Qwen3.8-Max, Qwen3.8-27B,
GLM-5.3 Flash, Qwen3.8-Flash, Gemini 3.8 Flash, GPT-6 Astra and DeepSeek V4.1 Flash. Reasoning models may use up to 64k output tokens; if the initial response is empty or unparsable, it is retried once at low reasoning effort. 

\paragraph{Metrics and statistics.}
Recall is pooled over lesions $\geq$1~mL and false positives are counted per patient; $J$ (Eq.~(\ref{eq:objective})) is computed over the
dense read's operating points. Uncertainty is a patient-level bootstrap with 2,000 resamples; false-positive differences use a paired
Wilcoxon signed-rank test. Loss attribution uses $G=0$ on development
and test patients and $G=8$ on training patients.

\paragraph{Self-improvement protocol.}
The designer (Claude Opus 5; GPT-6 Astra, Gemini 3.1 Pro and Gemini 3.8 Flash in Appendix~\ref{app:designers}) runs in a sandbox and
receives only aggregate trial statistics, attribution summaries and trial histories. Each recipe search trains 4--10
proposals for 100 steps after 2--3 seeds of the seed recipe set the noise floor, and retrains the leading proposals for 400 steps
before acceptance; for an accepted training revision, the median-$J$ seed becomes the next incumbent. The perception round trains six single-change proposals at trial scale on a
120-patient subset and replicates the two best; inference programs are revised up to eight times on cached development reads. The acquisition study is
reported as a separate component experiment, under the budgeted-read protocol with geometric screening only, because it was not submitted to
the full promotion gate.

\renewcommand{\thetable}{A\arabic{table}}\renewcommand{\thefigure}{A\arabic{figure}}\setcounter{table}{0}\setcounter{figure}{0}

\section{Frozen frontier readers}
\label{app:frozen}

\subsection{Does Better Acquisition Help Frozen Readers?}
% Refocused by hand (training-stage rows are in Appendix Table ablation_main; the dense-read block is in Table sizes_protocol).
\begin{table}[H]
\centering
\small
\setlength{\tabcolsep}{3.2pt}
\caption{\textbf{Baseline check: better views do not help frontier readers.} Liver, 31 held-out patients (111 lesions $\geq$1\,mL).
In each column every reader receives the same images and prompt, columns change only which slices are shown. R: recall with
patient-bootstrap 95\% interval (8-slice column); FP: false positives per patient.}
\label{tab:frozen_readers}
\begin{adjustbox}{max width=\linewidth}
\begin{tabular}{l cc cc cc}
\toprule
 & \multicolumn{2}{c}{8 uniform slices} & \multicolumn{2}{c}{Acquisition program} & \multicolumn{2}{c}{Lesion-guided (oracle)} \\
 & \multicolumn{2}{c}{\scriptsize (protocol of Table~\ref{tab:sizes})} & \multicolumn{2}{c}{\scriptsize (4 + up to 4 chosen views)} & \multicolumn{2}{c}{\scriptsize (slices through every lesion)} \\
\cmidrule(lr){2-3}\cmidrule(lr){4-5}\cmidrule(lr){6-7}
Reader & R & FP & R & FP & R & FP \\
\midrule
\multicolumn{7}{l}{\textit{Frozen frontier models, sorted by 8-slice recall}} \\
Gemini 3.8 Flash & \textbf{0.36} {\scriptsize[0.26, 0.44]} & 0.19 & 0.40 & 0.19 & 0.38 & 0.10 \\
Gemini 3.1 Pro & 0.35 {\scriptsize[0.24, 0.44]} & 1.13 & 0.38 & 0.90 & 0.35 & 1.10 \\
GPT-6 Astra & 0.32 {\scriptsize[0.20, 0.43]} & 0.58 & \textbf{0.42} & 0.39 & \textbf{0.43} & 0.55 \\
Qwen3.8-Max & 0.32 {\scriptsize[0.25, 0.38]} & 0.87 & 0.29 & 1.19 & 0.33 & 0.90 \\
Gemma 4 31B & 0.18 {\scriptsize[0.09, 0.31]} & 0.45 & 0.26 & 0.77 & 0.23 & 0.90 \\
GPT-5.5 & 0.15 {\scriptsize[0.09, 0.21]} & 0.58 & 0.14 & 0.87 & 0.19 & 1.00 \\
GLM-5.3 Flash & 0.15 {\scriptsize[0.08, 0.22]} & 1.26 & 0.23 & 2.03 & 0.24 & 1.97 \\
Mistral Large 3 & 0.12 {\scriptsize[0.06, 0.19]} & 5.19 & 0.10 & 4.97 & 0.13 & 4.16 \\
Qwen3.8-Flash & 0.11 {\scriptsize[0.05, 0.18]} & 1.52 & 0.14 & 1.42 & 0.20 & 1.84 \\
Claude Opus 5 & 0.10 {\scriptsize[0.05, 0.15]} & 1.29 & 0.14 & 1.68 & 0.13 & 1.68 \\
Qwen3.8-27B & 0.10 {\scriptsize[0.06, 0.14]} & 0.39 & 0.07 & 0.81 & 0.12 & 0.74 \\
Qwen3.5-397B & 0.07 {\scriptsize[0.02, 0.13]} & 1.23 & 0.13 & 1.39 & 0.13 & 1.19 \\
DeepSeek V4.1 Flash & 0.05 {\scriptsize[0.02, 0.10]} & 2.61 & 0.08 & 2.16 & 0.05 & 1.45 \\
\midrule
\multicolumn{7}{l}{\textit{Our Qwen3-VL-4B reader with a frozen vision encoder, on the same views and prompt}} \\
RL reader (supervised warm start + one RL round) & 0.27 {\scriptsize[0.17, 0.38]} & 0.97 & 0.33 & 0.61 & 0.37 & 0.35 \\
\bottomrule
\end{tabular}
\end{adjustbox}
\end{table}

To separate acquisition from perception, we evaluate thirteen frozen frontier readers and our frozen-encoder RL reader on the same 31 held-out liver patients while changing only the views they receive. We compare the standard 8-slice survey, views selected by the acquisition program, and oracle slices passing through every annotated lesion (Table~\ref{tab:frozen_readers}).

On the survey, frontier-model recall ranges from 0.05 to 0.36, with Gemini 3.8 Flash reaching 0.36 at 0.19 false positives per patient. Better exposure produces only modest gains. The acquisition program improves the strongest reader by at most 0.10 (GPT-6 Astra, 0.32 to 0.42), while even oracle lesion-guided slices improve it by only 0.11. Several readers change almost not at all: Gemini 3.8 Flash moves from 0.36 to 0.38, Qwen3.8-Max from 0.32 to 0.33, and DeepSeek V4.1 Flash remains at 0.05. Our frozen-encoder RL reader exhibits the same pattern (0.27, 0.33, and 0.37).

Thus, simply exposing the reader to more informative views does not close the detection gap. Under this protocol, the dominant limitation of frozen readers lies in recognizing visible evidence rather than only in selecting where to look, consistent with the attribution analysis in Section~\ref{sec:res-dynamics}.

\subsection{Lesion Size Reveals the Perception Gap}
% Auto-generated by make_all_tables.py from RESULTS_ALL.tex -- edit the data, not this file.
\begin{table}[H]
\centering
\small
\setlength{\tabcolsep}{3.4pt}
\caption{\textbf{Recall by lesion volume, 31 held-out liver patients.} Pooled over lesions; each cell is recall (hits/lesions). The same patients are the liver test set of the multi-organ study. Bold: best per column. The two largest bins hold 17 and 11 lesions, so one lesion moves recall by 0.06 and 0.09. Support: the number of slices on which a finding is reported (Appendix~\ref{app:terms}).}
\label{tab:sizes_full}
\begin{adjustbox}{max width=\linewidth}
\begin{tabular}{l cccc cc}
\toprule
Reader & 1--5\,mL & 5--20\,mL & 20--100\,mL & $>$100\,mL & All & FP \\
\textit{lesions} & \textit{n}=52 & \textit{n}=31 & \textit{n}=17 & \textit{n}=11 & \textit{n}=111 & \\
\midrule
\multicolumn{7}{l}{\textit{Frozen frontier readers, 8 uniform slices}} \\
Gemini 3.8 Flash & 0.08\,{\scriptsize(4/52)} & 0.48\,{\scriptsize(15/31)} & 0.65\,{\scriptsize(11/17)} & \textbf{0.91}\,{\scriptsize(10/11)} & 0.36 & 0.19 \\
Gemini 3.1 Pro & 0.12\,{\scriptsize(6/52)} & 0.42\,{\scriptsize(13/31)} & 0.71\,{\scriptsize(12/17)} & 0.73\,{\scriptsize(8/11)} & 0.35 & 1.13 \\
Qwen3.8-Max & 0.13\,{\scriptsize(7/52)} & 0.35\,{\scriptsize(11/31)} & 0.47\,{\scriptsize(8/17)} & \textbf{0.91}\,{\scriptsize(10/11)} & 0.32 & 0.87 \\
GPT-6 Astra & 0.13\,{\scriptsize(7/52)} & 0.29\,{\scriptsize(9/31)} & 0.65\,{\scriptsize(11/17)} & 0.82\,{\scriptsize(9/11)} & 0.32 & 0.58 \\
Gemma 4 31B & 0.02\,{\scriptsize(1/52)} & 0.10\,{\scriptsize(3/31)} & 0.41\,{\scriptsize(7/17)} & 0.82\,{\scriptsize(9/11)} & 0.18 & 0.45 \\
GPT-5.5 & 0.04\,{\scriptsize(2/52)} & 0.13\,{\scriptsize(4/31)} & 0.24\,{\scriptsize(4/17)} & 0.64\,{\scriptsize(7/11)} & 0.15 & 0.58 \\
GLM-5.3 Flash & 0.04\,{\scriptsize(2/52)} & 0.10\,{\scriptsize(3/31)} & 0.29\,{\scriptsize(5/17)} & 0.64\,{\scriptsize(7/11)} & 0.15 & 1.26 \\
Mistral Large 3 & 0.06\,{\scriptsize(3/52)} & 0.00\,{\scriptsize(0/31)} & 0.06\,{\scriptsize(1/17)} & 0.82\,{\scriptsize(9/11)} & 0.12 & 5.19 \\
Qwen3.8-Flash & 0.02\,{\scriptsize(1/52)} & 0.10\,{\scriptsize(3/31)} & 0.18\,{\scriptsize(3/17)} & 0.45\,{\scriptsize(5/11)} & 0.11 & 1.52 \\
Claude Opus 5 & 0.00\,{\scriptsize(0/52)} & 0.10\,{\scriptsize(3/31)} & 0.18\,{\scriptsize(3/17)} & 0.45\,{\scriptsize(5/11)} & 0.10 & 1.29 \\
Qwen3.8-27B & 0.00\,{\scriptsize(0/52)} & 0.03\,{\scriptsize(1/31)} & 0.24\,{\scriptsize(4/17)} & 0.55\,{\scriptsize(6/11)} & 0.10 & 0.39 \\
Qwen3.5-397B & 0.00\,{\scriptsize(0/52)} & 0.10\,{\scriptsize(3/31)} & 0.06\,{\scriptsize(1/17)} & 0.36\,{\scriptsize(4/11)} & 0.07 & 1.23 \\
DeepSeek V4.1 Flash & 0.00\,{\scriptsize(0/52)} & 0.00\,{\scriptsize(0/31)} & 0.00\,{\scriptsize(0/17)} & 0.55\,{\scriptsize(6/11)} & 0.05 & 2.61 \\
Qwen3-VL-4B base & 0.02\,{\scriptsize(1/52)} & 0.03\,{\scriptsize(1/31)} & 0.06\,{\scriptsize(1/17)} & 0.18\,{\scriptsize(2/11)} & 0.05 & 1.32 \\
\midrule
\multicolumn{7}{l}{\textit{Ours, the same 8 uniform slices}} \\
\textit{Warm start} & 0.08\,{\scriptsize(4/52)} & 0.23\,{\scriptsize(7/31)} & 0.47\,{\scriptsize(8/17)} & 0.82\,{\scriptsize(9/11)} & 0.25 & 0.42 \\
\textit{RL reader} & 0.08\,{\scriptsize(4/52)} & 0.26\,{\scriptsize(8/31)} & 0.53\,{\scriptsize(9/17)} & 0.82\,{\scriptsize(9/11)} & 0.27 & 0.97 \\
\textit{Fresh-view reader} & 0.08\,{\scriptsize(4/52)} & 0.26\,{\scriptsize(8/31)} & 0.71\,{\scriptsize(12/17)} & \textbf{0.91}\,{\scriptsize(10/11)} & 0.31 & 0.68 \\
\textit{Second fresh-view reader} & 0.08\,{\scriptsize(4/52)} & 0.26\,{\scriptsize(8/31)} & 0.71\,{\scriptsize(12/17)} & 0.82\,{\scriptsize(9/11)} & 0.30 & 0.29 \\
\midrule
\multicolumn{7}{l}{\textit{Ours, every slice with 3D reconciliation (support $\geq$2)}} \\
\textit{RL reader} & 0.29\,{\scriptsize(15/52)} & 0.39\,{\scriptsize(12/31)} & 0.76\,{\scriptsize(13/17)} & \textbf{0.91}\,{\scriptsize(10/11)} & 0.45 & 0.52 \\
\textit{Fresh-view reader} & 0.31\,{\scriptsize(16/52)} & 0.45\,{\scriptsize(14/31)} & 0.82\,{\scriptsize(14/17)} & \textbf{0.91}\,{\scriptsize(10/11)} & 0.49 & 0.42 \\
\textit{Second fresh-view reader} & 0.25\,{\scriptsize(13/52)} & 0.48\,{\scriptsize(15/31)} & 0.76\,{\scriptsize(13/17)} & 0.82\,{\scriptsize(9/11)} & 0.45 & 0.32 \\
\textit{Cascade: RL proposes, cached-view confirms} & 0.37\,{\scriptsize(19/52)} & 0.58\,{\scriptsize(18/31)} & 0.88\,{\scriptsize(15/17)} & \textbf{0.91}\,{\scriptsize(10/11)} & 0.56 & 0.61 \\
\midrule
\multicolumn{7}{l}{\textit{Frozen reader under our protocol: dense read (every slice), $\leq$0.5 FP per patient}} \\
Gemini 3.8 Flash, every slice & 0.23\,{\scriptsize(12/52)} & 0.32\,{\scriptsize(10/31)} & 0.41\,{\scriptsize(7/17)} & 0.82\,{\scriptsize(9/11)} & 0.34 & 0.26 \\
\midrule
\multicolumn{7}{l}{\textit{Ours, final multi-organ reader, every slice, $\leq$0.5 FP per patient (test evaluation)}} \\
\textit{Final reader (designer's configuration, $s_0$)} & \textbf{0.69}\,{\scriptsize(36/52)} & \textbf{0.84}\,{\scriptsize(26/31)} & \textbf{1.00}\,{\scriptsize(17/17)} & 0.82\,{\scriptsize(9/11)} & \textbf{0.79} & 0.35 \\
\bottomrule
\end{tabular}
\end{adjustbox}
\end{table}

Table~\ref{tab:sizes_full} stratifies lesion recall by volume for every reader on the 31 held-out liver patients. Lesion size strongly determines performance: below 5\,mL, no frontier model detects more than 0.13 of lesions (7 of 52), whereas ten of thirteen detect at least 0.55 of lesions above 100\,mL.

Training the language-side reader improves larger lesions without removing this small-lesion ceiling. On the survey, our readers increase 20--100\,mL recall from 0.06 before training to 0.71 after the fresh-view recipe, while 1--5\,mL recall remains between 0.02 and 0.08. Dense reading improves the smallest lesions to 0.29 for the RL reader, 0.31 for the fresh-view reader, and 0.37 for the cascade, but the largest change follows perception adaptation. The final reader detects 36 of 52 lesions of 1--5\,mL (0.69), 26 of 31 lesions of 5--20\,mL, and all 17 lesions of 20--100\,mL.

The size-stratified results therefore localize the remaining difficulty: the largest gains from changing perception occur precisely in the small-lesion regime that frozen readers and recipe-level revisions fail to resolve. Because individual bins contain 11--52 lesions, one lesion changes a reported recall by 0.02--0.09.

\subsection{Separating Reading Protocol from Perception}
% Auto-generated by make_all_tables.py from RESULTS_ALL.tex -- edit the data, not this file.
\begin{table}[H]
\centering
\footnotesize
\setlength{\tabcolsep}{2.8pt}
\caption{\textbf{Reading protocol and perception configuration, by lesion volume} (test patients). Recall at $\leq$0.5 false positives per patient; FP: false positives per patient at that operating point. Reading every slice does not help the frozen frontier model at a comparable false-positive rate (0.36 on the survey, 0.34 dense on liver), whereas adapting the vision encoder raises our reader's recall in every size bin below 100\,mL. Bold: best in column.}
\label{tab:sizes_protocol}
\begin{adjustbox}{max width=\linewidth}
\begin{tabular}{l cccccc cccccc}
\toprule
 & \multicolumn{6}{c}{Liver: 31 patients, 111 lesions} & \multicolumn{6}{c}{Kidney: 70 patients, 81 lesions} \\
\cmidrule(lr){2-7}\cmidrule(lr){8-13}
Reader & 1--5 & 5--20 & 20--100 & $>$100 & All & FP & 1--5 & 5--20 & 20--100 & $>$100 & All & FP \\
\midrule
\multicolumn{13}{l}{\textit{Strongest frontier model: survey vs.\ our protocol}} \\
Gemini 3.8 Flash, 8 uniform slices & 0.08 & 0.48 & 0.65 & \textbf{0.91} & 0.36 & 0.19 & 0.21 & 0.46 & 0.68 & \textbf{1.00} & 0.60 & 0.13 \\
Gemini 3.8 Flash, dense read (our protocol) & 0.23 & 0.32 & 0.41 & 0.82 & 0.34 & 0.26 & 0.07 & 0.25 & 0.64 & 0.72 & 0.44 & 0.13 \\
\midrule
\multicolumn{13}{l}{\textit{Ours: one 4B reader, dense read; perception configurations}} \\
\textit{Frozen vision encoder} & 0.06 & 0.13 & 0.41 & 0.64 & 0.19 & 0.00 & 0.14 & 0.29 & 0.88 & \textbf{1.00} & 0.60 & 0.50 \\
\textit{Adapters on 4 blocks (human-chosen)} & 0.12 & 0.19 & 0.53 & 0.64 & 0.25 & 0.06 & 0.29 & 0.50 & 0.92 & \textbf{1.00} & 0.70 & 0.46 \\
\textit{ReVision3D (24 blocks, designer)} & \textbf{0.69} & \textbf{0.84} & \textbf{1.00} & 0.82 & \textbf{0.79} & 0.35 & \textbf{0.57} & \textbf{0.71} & \textbf{0.96} & \textbf{1.00} & \textbf{0.83} & 0.37 \\
\bottomrule
\end{tabular}
\end{adjustbox}
\end{table}

Table~\ref{tab:sizes_protocol} controls for the reading protocol by evaluating the strongest frontier reader under both the survey and our dense read, then comparing our three perception configurations under the same dense protocol.

Dense reading alone does not explain the gain. For Gemini 3.8 Flash, it improves only the smallest liver-lesion bin (0.08 to 0.23) while reducing recall on larger lesions; overall liver recall remains essentially unchanged (0.36 to 0.34), and kidney recall falls from 0.60 to 0.44. Additional slices also introduce more false positives, forcing a stricter support threshold at the target operating point. In contrast, under the identical dense protocol, our liver reader moves from 0.19 with a frozen encoder to 0.25 with the hand-selected adapters and 0.79 with the designer-selected perception configuration. Kidney recall similarly rises from 0.60 to 0.70 and 0.83.

The advantage of ReVision3D therefore cannot be attributed to reading more slices alone. Holding the protocol fixed, the dominant improvement comes from changing the visual representation itself.

\subsection{Kidney Provides a Higher-Contrast Test Case}
% Edited by hand: our block shows the dense read (the reader's operating mode).
\begin{table}[H]
\centering
\footnotesize
\setlength{\tabcolsep}{2.8pt}
\caption{\textbf{Baseline check on kidney: frontier readers on the survey and on lesion-guided slices}}
\label{tab:readers_kidney}
\begin{adjustbox}{max width=\linewidth}
\begin{tabular}{l cc cc}
\toprule
 & \multicolumn{2}{c}{8 uniform slices} & \multicolumn{2}{c}{Lesion-guided (oracle)} \\
\cmidrule(lr){2-3}\cmidrule(lr){4-5}
Reader & R & FP & R & FP \\
\midrule
\multicolumn{5}{l}{\textit{Frozen frontier models, in order of release (bold: best per column)}} \\
Mistral Large 3 & 0.15 {\scriptsize[0.08, 0.23]} & 2.04 & 0.15 & 1.93 \\
Qwen3.5-397B & 0.41 {\scriptsize[0.30, 0.54]} & 0.81 & 0.46 & 0.80 \\
Gemini 3.1 Pro & 0.52 {\scriptsize[0.40, 0.66]} & 0.43 & 0.67 & 0.24 \\
Gemma 4 31B & 0.54 {\scriptsize[0.44, 0.66]} & 0.26 & 0.59 & 0.27 \\
GPT-5.5 & 0.42 {\scriptsize[0.30, 0.56]} & 0.20 & 0.53 & 0.26 \\
Claude Opus 5 & 0.26 {\scriptsize[0.16, 0.37]} & 0.77 & 0.27 & 0.86 \\
Qwen3.8-Max & 0.46 {\scriptsize[0.34, 0.59]} & 0.24 & 0.53 & 0.26 \\
Qwen3.8-27B & 0.31 {\scriptsize[0.21, 0.42]} & 0.50 & 0.51 & 0.33 \\
GLM-5.3 Flash & 0.37 {\scriptsize[0.28, 0.48]} & 0.83 & 0.47 & 0.83 \\
Qwen3.8-Flash & 0.42 {\scriptsize[0.32, 0.54]} & 0.64 & 0.41 & 0.86 \\
Gemini 3.8 Flash & 0.60 {\scriptsize[0.47, 0.75]} & 0.13 & 0.74 & 0.06 \\
GPT-6 Astra & \textbf{0.70} {\scriptsize[0.57, 0.85]} & 0.14 & \textbf{0.79} & 0.09 \\
DeepSeek V4.1 Flash & 0.25 {\scriptsize[0.14, 0.36]} & 0.87 & 0.28 & 0.80 \\
\midrule
\multicolumn{5}{l}{\textit{Ours, same 70 patients, dense read (every slice)}} \\
Final reader, support $\geq$2 & \multicolumn{4}{l}{0.72 {\scriptsize[0.59, 0.85]} at 0.07 FP per patient} \\
Final reader, support $\geq$1 & \multicolumn{4}{l}{\textbf{0.83} {\scriptsize[0.73, 0.93]} at 0.37 FP per patient} \\
Final reader + inference program & \multicolumn{4}{l}{0.79 / 0.81 / 0.85 within 0.25 / 0.5 / 1 FP} \\
\bottomrule
\end{tabular}
\end{adjustbox}
\end{table}

% Auto-generated by make_all_tables.py from RESULTS_ALL.tex -- edit the data, not this file.
\begin{table}[H]
\centering
\small
\setlength{\tabcolsep}{3.4pt}
\caption{\textbf{Recall by tumour volume, 70 kidney test patients.} Pooled over tumours; each cell is recall (hits/tumours). Support $\geq$1 is the $\leq$0.5 FP operating point used in Table~\ref{tab:sizes}.}
\label{tab:sizes_kidney_full}
\begin{adjustbox}{max width=\linewidth}
\begin{tabular}{l cccc cc}
\toprule
Reader & 1--5\,mL & 5--20\,mL & 20--100\,mL & $>$100\,mL & All & FP \\
\textit{tumours} & \textit{n}=14 & \textit{n}=24 & \textit{n}=25 & \textit{n}=18 & \textit{n}=81 & \\
\midrule
\multicolumn{7}{l}{\textit{Frozen frontier models, 8 uniform slices, in order of release}} \\
Mistral Large 3 & 0.00\,{\scriptsize(0/14)} & 0.04\,{\scriptsize(1/24)} & 0.00\,{\scriptsize(0/25)} & 0.61\,{\scriptsize(11/18)} & 0.15 & 2.04 \\
Qwen3.5-397B & 0.00\,{\scriptsize(0/14)} & 0.08\,{\scriptsize(2/24)} & 0.56\,{\scriptsize(14/25)} & 0.94\,{\scriptsize(17/18)} & 0.41 & 0.81 \\
Gemini 3.1 Pro & 0.21\,{\scriptsize(3/14)} & 0.17\,{\scriptsize(4/24)} & 0.68\,{\scriptsize(17/25)} & \textbf{1.00}\,{\scriptsize(18/18)} & 0.52 & 0.43 \\
Gemma 4 31B & 0.14\,{\scriptsize(2/14)} & 0.29\,{\scriptsize(7/24)} & 0.72\,{\scriptsize(18/25)} & 0.94\,{\scriptsize(17/18)} & 0.54 & 0.26 \\
GPT-5.5 & 0.00\,{\scriptsize(0/14)} & 0.17\,{\scriptsize(4/24)} & 0.60\,{\scriptsize(15/25)} & 0.83\,{\scriptsize(15/18)} & 0.42 & 0.20 \\
Claude Opus 5 & 0.00\,{\scriptsize(0/14)} & 0.00\,{\scriptsize(0/24)} & 0.28\,{\scriptsize(7/25)} & 0.78\,{\scriptsize(14/18)} & 0.26 & 0.77 \\
Qwen3.8-Max & 0.14\,{\scriptsize(2/14)} & 0.17\,{\scriptsize(4/24)} & 0.52\,{\scriptsize(13/25)} & \textbf{1.00}\,{\scriptsize(18/18)} & 0.46 & 0.24 \\
Qwen3.8-27B & 0.00\,{\scriptsize(0/14)} & 0.04\,{\scriptsize(1/24)} & 0.36\,{\scriptsize(9/25)} & 0.83\,{\scriptsize(15/18)} & 0.31 & 0.50 \\
Qwen3.8-Flash & 0.00\,{\scriptsize(0/14)} & 0.21\,{\scriptsize(5/24)} & 0.48\,{\scriptsize(12/25)} & 0.94\,{\scriptsize(17/18)} & 0.42 & 0.64 \\
GLM-5.3 Flash & 0.07\,{\scriptsize(1/14)} & 0.21\,{\scriptsize(5/24)} & 0.32\,{\scriptsize(8/25)} & 0.89\,{\scriptsize(16/18)} & 0.37 & 0.83 \\
Gemini 3.8 Flash & 0.21\,{\scriptsize(3/14)} & 0.46\,{\scriptsize(11/24)} & 0.68\,{\scriptsize(17/25)} & \textbf{1.00}\,{\scriptsize(18/18)} & 0.60 & 0.13 \\
GPT-6 Astra & 0.14\,{\scriptsize(2/14)} & 0.58\,{\scriptsize(14/24)} & 0.92\,{\scriptsize(23/25)} & \textbf{1.00}\,{\scriptsize(18/18)} & 0.70 & 0.14 \\
DeepSeek V4.1 Flash & 0.07\,{\scriptsize(1/14)} & 0.04\,{\scriptsize(1/24)} & 0.24\,{\scriptsize(6/25)} & 0.67\,{\scriptsize(12/18)} & 0.25 & 0.87 \\
\midrule
\multicolumn{7}{l}{\textit{Frozen model under our protocol: dense read, $\leq$0.5 FP per patient}} \\
Gemini 3.8 Flash, every slice & 0.07\,{\scriptsize(1/14)} & 0.25\,{\scriptsize(6/24)} & 0.64\,{\scriptsize(16/25)} & 0.72\,{\scriptsize(13/18)} & 0.44 & 0.13 \\
\midrule
\multicolumn{7}{l}{\textit{Ours, final reader, dense read}} \\
\textit{Final reader, dense read, support $\geq$2} & 0.29\,{\scriptsize(4/14)} & 0.58\,{\scriptsize(14/24)} & 0.88\,{\scriptsize(22/25)} & \textbf{1.00}\,{\scriptsize(18/18)} & 0.72 & 0.07 \\
\textit{Final reader, dense read, support $\geq$1} & \textbf{0.57}\,{\scriptsize(8/14)} & \textbf{0.71}\,{\scriptsize(17/24)} & \textbf{0.96}\,{\scriptsize(24/25)} & \textbf{1.00}\,{\scriptsize(18/18)} & \textbf{0.83} & 0.37 \\
\bottomrule
\end{tabular}
\end{adjustbox}
\end{table}

We repeat the frozen-reader comparison on 70 kidney test patients containing 81 tumors $\geq1$\,mL (Tables~\ref{tab:readers_kidney} and~\ref{tab:sizes_kidney_full}). Kidney tumors are generally larger and higher contrast than liver lesions, and the frontier readers accordingly perform substantially better: survey recall ranges from 0.15 to 0.70, led by GPT-6 Astra at 0.70 recall and 0.14 false positives per patient.

Oracle exposure has a larger effect than on liver for some readers. GPT-6 Astra reaches 0.79 and Gemini 3.1 Pro rises from 0.52 to 0.67, but weak readers again change little. ReVision3D reaches 0.72 recall at 0.07 false positives per patient and 0.83 at 0.37; after inference-level revision it reaches 0.79, 0.81, and 0.85 recall within false-positive budgets of 0.25, 0.5, and 1 per patient.

The remaining difference again concentrates in small tumors. Below 5\,mL, no frontier model detects more than 3 of 14 tumors (0.21), whereas our reader detects 8 (0.57). Above 20\,mL, the strongest systems approach ceiling. Kidney therefore provides a complementary setting in which frontier readers are already competitive on large, conspicuous tumors, while the perception advantage remains visible in the smaller-lesion regime.

\section{The final reader: perception level, liver and kidney}
\label{app:final}

\subsection{Attribution Across Perception Configurations}
% Auto-generated by make_all_tables.py from RESULTS_ALL.tex -- edit the data, not this file.
\begin{table}[H]
\centering
\small
\setlength{\tabcolsep}{4pt}
\caption{\textbf{Where lesions are lost} (Eq.~\eqref{eq:attribution} with $G=0$) at 0.5 false positives per patient, development patients. Each row splits the lesions $\geq$1\,mL into those no slice read detects, those detected and then discarded by aggregation, and those kept (recall); rows sum to 1. Bold: lowest never-detected and highest kept share.}
\label{tab:multi_attribution}
\begin{adjustbox}{max width=\linewidth}
\begin{tabular}{l ccc ccc}
\toprule
 & \multicolumn{3}{c}{Liver} & \multicolumn{3}{c}{Kidney} \\
\cmidrule(lr){2-4}\cmidrule(lr){5-7}
Perception configuration & Never detected & Discarded & Kept & Never detected & Discarded & Kept \\
\midrule
Frozen encoder & 0.42 & 0.13 & 0.45 & 0.45 & 0.00 & 0.55 \\
Adapters, hand & 0.30 & 0.20 & 0.50 & 0.30 & 0.00 & 0.70 \\
Adapters, designer & \textbf{0.25} & 0.05 & \textbf{0.70} & \textbf{0.14} & 0.00 & \textbf{0.86} \\
\bottomrule
\end{tabular}
\end{adjustbox}
\end{table}

Table~\ref{tab:multi_attribution} applies the failure decomposition of Eq.~(\ref{eq:attribution}) to the three perception configurations on development patients. Each lesion is classified as never detected, detected but discarded, or retained in the final prediction.

On liver, adapting perception progressively reduces the never-detected share from 0.42 with the frozen encoder to 0.30 with the hand-selected adapters and 0.25 with the designer-selected configuration. The corresponding kept share rises from 0.45 to 0.50 and 0.70. The hand-selected configuration illustrates why recall alone does not identify the responsible mechanism: it detects additional lesions but also discards more of them (0.13 to 0.20), yielding only a small net improvement. The designer-selected configuration reduces both sources of loss, with the discarded share falling to 0.05. On kidney, no detected tumor is discarded, so the entire gain comes from reducing the never-detected share from 0.45 to 0.30 and finally 0.14.

These decompositions show that perception adaptation changes the loss component it is expected to affect: previously unseen findings become detectable, rather than the gain arising only from a different aggregation threshold.

\subsection{Perception Gains Persist Across Lesion Sizes}
% Auto-generated by make_all_tables.py from RESULTS_ALL.tex -- edit the data, not this file.
\begin{table}[H]
\centering
\small
\setlength{\tabcolsep}{4pt}
\caption{Recall by lesion volume on the development patients (seed $s_0$), at the most permissive support threshold with at most 0.5 false positives per patient. The test-set counterpart is Table~\ref{tab:sizes}. Support: the number of slices on which a finding is reported (Appendix~\ref{app:terms}).}
\label{tab:multi_sizes}
\begin{adjustbox}{max width=\linewidth}
\begin{tabular}{l cccc cc}
\toprule
Perception configuration & 1--5\,mL & 5--20\,mL & 20--100\,mL & $>$100\,mL & FP & P \\
\midrule
\multicolumn{7}{l}{\textit{Liver (1--5: 26, 5--20: 15, 20--100: 16, $>$100: 3 lesions)}} \\
Frozen encoder & 0.19\,{\scriptsize(5/26)} & 0.60\,{\scriptsize(9/15)} & 0.62\,{\scriptsize(10/16)} & \textbf{1.00}\,{\scriptsize(3/3)} & 0.20 & 0.93 \\
Adapters, hand & 0.35\,{\scriptsize(9/26)} & 0.53\,{\scriptsize(8/15)} & 0.62\,{\scriptsize(10/16)} & \textbf{1.00}\,{\scriptsize(3/3)} & 0.30 & 0.92 \\
Adapters, designer & \textbf{0.50}\,{\scriptsize(13/26)} & \textbf{0.73}\,{\scriptsize(11/15)} & \textbf{0.94}\,{\scriptsize(15/16)} & \textbf{1.00}\,{\scriptsize(3/3)} & 0.05 & 0.99 \\
\midrule
\multicolumn{7}{l}{\textit{Kidney (1--5: 10, 5--20: 10, 20--100: 13, $>$100: 11 lesions)}} \\
Frozen encoder & 0.10\,{\scriptsize(1/10)} & 0.40\,{\scriptsize(4/10)} & 0.62\,{\scriptsize(8/13)} & \textbf{1.00}\,{\scriptsize(11/11)} & 0.50 & 0.83 \\
Adapters, hand & 0.40\,{\scriptsize(4/10)} & 0.60\,{\scriptsize(6/10)} & 0.77\,{\scriptsize(10/13)} & \textbf{1.00}\,{\scriptsize(11/11)} & 0.38 & 0.88 \\
Adapters, designer & \textbf{0.60}\,{\scriptsize(6/10)} & \textbf{0.80}\,{\scriptsize(8/10)} & \textbf{1.00}\,{\scriptsize(13/13)} & \textbf{1.00}\,{\scriptsize(11/11)} & 0.25 & 0.92 \\
\bottomrule
\end{tabular}
\end{adjustbox}
\end{table}

Table~\ref{tab:multi_sizes} reports the three perception configurations on the development patients used for selection, stratified by lesion volume. The designer-selected configuration performs best in every sub-100\,mL bin on both organs while also reducing false positives.

On liver, recall for 1--5\,mL lesions increases from 0.19 to 0.35 and 0.50 across the frozen, hand-selected, and designer-selected configurations, while recall for 20--100\,mL lesions rises from 0.62 to 0.94. On kidney, 1--5\,mL recall increases from 0.10 to 0.40 and 0.60, while false positives fall from 0.50 to 0.25 per patient. The largest-volume bins contain few lesions and saturate across configurations.

The ordering observed on held-out test patients is therefore already present on the development cohort used for selection. Although all configurations exhibit some development-to-test drop in $J$, the relative advantage of the designer-selected perception configuration is consistent across organs and lesion sizes.

\subsection{Searching the Perception Level}
% Auto-generated by make_all_tables.py from RESULTS_ALL.tex -- edit the data, not this file.
\begin{table}[H]
\centering
\small
\setlength{\tabcolsep}{3pt}
\caption{\textbf{Perception-level round at trial scale} (120 training patients; 30 liver and kidney development patients). Each proposal changes one setting of the vision-encoder adaptation. $\bar J$ averages liver and kidney; $\Delta$ is the difference from the mean of the two default seeds (0.381), whose range, 0.023, is the noise floor. The selected configuration averages 0.678 over two seeds and was accepted.}
\label{tab:perception_round}
\begin{adjustbox}{max width=\linewidth}
\begin{tabular}{ll l c c cc}
\toprule
Trial & Change & Configuration & $\bar J$ & $\Delta$ & Liver $J$ (never det.) & Kidney $J$ (never det.) \\
\midrule
base\_s0 & default & 4 blocks, r16, 1.0$\times$, 1 ep & 0.369 & -- & 0.293 {\scriptsize(0.62)} & 0.446 {\scriptsize(0.52)} \\
base\_s1 & default (seed 2) & 4 blocks, r16, 1.0$\times$, 1 ep & 0.392 & -- & 0.328 {\scriptsize(0.52)} & 0.457 {\scriptsize(0.52)} \\
\midrule
p01 & 2$\times$ views & 4 blocks, r16, 2.0$\times$, 1 ep & 0.403 & +0.023 & 0.328 {\scriptsize(0.45)} & 0.478 {\scriptsize(0.48)} \\
p02 & 12 blocks & 12 blocks, r16, 2.0$\times$, 1 ep & 0.496 & +0.115 & 0.491 {\scriptsize(0.48)} & 0.500 {\scriptsize(0.48)} \\
p03 & 24 blocks & 24 blocks, r16, 2.0$\times$, 1 ep & 0.535 & +0.155 & 0.440 {\scriptsize(0.24)} & 0.630 {\scriptsize(0.17)} \\
p04 & rank 32 & 24 blocks, r32, 2.0$\times$, 1 ep & 0.484 & +0.103 & 0.457 {\scriptsize(0.17)} & 0.511 {\scriptsize(0.48)} \\
p05 & 2 epochs & 24 blocks, r32, 2.0$\times$, 2 ep & 0.716 & +0.335 & 0.716 {\scriptsize(0.14)} & 0.717 {\scriptsize(0.26)} \\
p06 & lr $3{\times}10^{-4}$ & 24 blocks, r32, 2.0$\times$, 2 ep, lr $3{\times}10^{-4}$ & 0.000 & diverged & 0.000 {\scriptsize(1.00)} & 0.000 {\scriptsize(1.00)} \\
p05\_s1 & p05, seed 2 & 24 blocks, r32, 2.0$\times$, 2 ep & 0.640 & +0.260 & 0.716 {\scriptsize(0.14)} & 0.565 {\scriptsize(0.39)} \\
p04\_s1 & p04, seed 2 & 24 blocks, r32, 2.0$\times$, 1 ep & 0.589 & +0.208 & 0.612 {\scriptsize(0.21)} & 0.565 {\scriptsize(0.43)} \\
\bottomrule
\end{tabular}
\end{adjustbox}
\end{table}

Table~\ref{tab:perception_round} records the perception-level search. Two runs of the default configuration first establish a trial-scale noise floor of 0.023; subsequent proposals change one perception setting at a time, and the two strongest proposals are independently repeated.

Increasing image resolution alone produces a gain equal to the noise floor and is therefore insufficient. In contrast, adapting 12 and then all 24 vision blocks increases $\bar J$ by 0.115 and 0.155, accompanied by large reductions in the never-detected share. Increasing adapter rank alone provides little additional benefit, whereas a second training epoch on top of the stronger adaptation produces the largest trial-scale objective, $\bar J=0.716$. Increasing the learning rate causes divergence. Replicates of the two leading configurations preserve their ordering, although their variation is larger than that of the default runs.

The search therefore identifies a qualitatively different intervention from additional views or recipe changes: exposing more of the vision encoder to adaptation, followed by longer training, directly reduces the never-detected component that motivated the perception-level search.

\section{Recipe level: trajectory, recipe searches and learnability}
\label{app:recipe}

\subsection{Accepted Revisions Change Different Parts of the Operating Curve}
% Auto-generated by make_all_tables.py from RESULTS_ALL.tex -- edit the data, not this file.
\begin{table}[H]
\centering
\small
\setlength{\tabcolsep}{3.4pt}
\caption{\textbf{The liver system after each accepted revision} (20 development patients, 60 lesions $\geq$1\,mL, dense read). The warm start is supervised on mask-derived reports; RL uses the mask-grounded reward. Max R is the recall reachable at any budget: it stays near 0.58 until the perception level is revised.}
\label{tab:trajectory}
\begin{adjustbox}{max width=\linewidth}
\begin{tabular}{ll cc cc c}
\toprule
 & & & & \multicolumn{2}{c}{R / P at FP per patient} & Max \\
\cmidrule(lr){5-6}
System after each stage & Change made by & $J$ & $\Delta J$ & $\leq$0.25 & $\leq$0.5 & R \\
\midrule
Warm start & fixed pipeline & 0.379 & -- & 0.17 / 0.93 & 0.45 / 0.88 & 0.58 \\
+\,RL & fixed pipeline & 0.463 & +0.084 & 0.30 / 1.00 & 0.52 / 0.91 & 0.58 \\
+\,learnability recipe & designer, 3 seeds & 0.529 & +0.066 & 0.42 / 1.00 & 0.57 / 0.92 & 0.57 \\
+\,vision-encoder adapters & hand, on designer's diagnosis & 0.575 & +0.046 & 0.53 / 0.95 & 0.53 / 0.95 & \textbf{0.70} \\
+\,inference program (own readers) & designer (accepted) & \textbf{0.608} & +0.033 & 0.58 / 0.97 & 0.58 / 0.97 & 0.68 \\
\bottomrule
\end{tabular}
\end{adjustbox}
\end{table}

Table~\ref{tab:trajectory} reports the liver system after each accepted revision. Across the trajectory, $J$ increases from 0.379 to 0.463, 0.529, 0.575, and finally 0.608. However, these gains arise through different mechanisms.

The early training stages primarily improve precision. Recall at 0.25 false positives per patient increases from 0.17 to 0.42, while the maximum recall attainable at any operating point remains approximately unchanged at 0.57--0.58. The perception revision is the first intervention that raises this ceiling, from 0.57 to 0.70. Precision remains at least 0.88 throughout.

Thus, an increasing scalar objective alone does not reveal what has improved. Recipe-level revisions make better use of detections already available to the reader, whereas the perception revision expands the set of lesions that the system can detect.

\subsection{First Recipe Search: Short-Run Gains Do Not Survive Promotion}

% Auto-generated by make_all_tables.py from RESULTS_ALL.tex -- edit the data, not this file.
\begin{table}[H]
\centering
\small
\setlength{\tabcolsep}{3pt}
\caption{
\textbf{First recipe-search round from the RL reader.}
Ten designer recipes are screened for 100 steps and the two strongest are
retrained for 400 steps. Two seed repeats define the 0.050 noise floor.
The final columns compare the measured and replay-predicted fractions of
training steps with learning signal.
}
\label{tab:recipe_roundA}
\begin{adjustbox}{max width=\linewidth}
\begin{tabular}{ll c ccc cc cc}
\toprule
 & & & \multicolumn{3}{c}{Gain} & \multicolumn{2}{c}{R / FP at support} & \multicolumn{2}{c}{Steps with gradient} \\
\cmidrule(lr){4-6}\cmidrule(lr){7-8}\cmidrule(lr){9-10}
Trial & Recipe by & $J$ & vs.\ start & $P(\Delta J{>}0)$ & vs.\ seed & $\geq$1 & $\geq$2 & measured & replay \\
\midrule
seed $s_0$ & seed recipe & 0.467 & +0.004 & 0.09 & -- & 0.60 / 3.25 & 0.47 / 0.25 & 0.29 & 0.36 \\
seed $s_1$ & seed recipe & 0.517 & +0.054 & 0.75 & -- & 0.58 / 0.95 & 0.45 / 0.05 & 0.34 & 0.36 \\
\midrule
p01 & designer & 0.533 & +0.071 & 0.88 & +0.041 & 0.58 / 0.35 & 0.38 / 0.00 & 0.55 & 0.76 \\
p02 & designer & 0.517 & +0.054 & 0.70 & +0.025 & 0.57 / 1.00 & 0.47 / 0.10 & 0.53 & 0.74 \\
p03 & designer & 0.500 & +0.037 & 0.55 & +0.008 & 0.58 / 0.90 & 0.42 / 0.10 & 0.40 & 0.83 \\
p04 & designer & 0.525 & +0.062 & 0.89 & +0.033 & 0.57 / 0.35 & 0.40 / 0.00 & 0.37 & 0.83 \\
p05 & designer & 0.500 & +0.037 & 0.66 & +0.008 & 0.58 / 0.80 & 0.42 / 0.05 & 0.66 & 0.84 \\
p06 & designer & 0.517 & +0.054 & 0.84 & +0.025 & 0.58 / 0.70 & 0.45 / 0.05 & 0.66 & 0.83 \\
p07 & designer & 0.525 & +0.062 & 0.82 & +0.033 & 0.58 / 0.85 & 0.47 / 0.05 & 0.50 & 0.82 \\
p08 & designer & 0.508 & +0.046 & 0.67 & +0.016 & 0.58 / 0.95 & 0.43 / 0.10 & 0.65 & 0.79 \\
p09 & designer & 0.508 & +0.046 & 0.72 & +0.016 & 0.58 / 0.85 & 0.43 / 0.10 & 0.50 & 0.77 \\
p10 & designer & 0.483 & +0.021 & 0.71 & $-$0.009 & 0.57 / 0.60 & 0.40 / 0.05 & 0.40 & 0.80 \\
p01, 400 steps & designer & 0.508 & +0.046 & 0.65 & +0.016 & 0.58 / 1.30 & 0.48 / 0.20 & 0.34 & 0.76 \\
p07, 400 steps & designer & 0.479 & +0.017 & 0.37 & $-$0.013 & 0.62 / 1.80 & 0.43 / 0.10 & 0.31 & 0.82 \\
hand$^\ast$, 400 steps & hand-designed & 0.529 & +0.067 & 0.86 & +0.037 & 0.57 / 0.40 & 0.42 / 0.00 & -- & -- \\
\bottomrule
\end{tabular}
\end{adjustbox}
\end{table}

\begin{figure}[H]
\centering
\includegraphics[width=\linewidth]{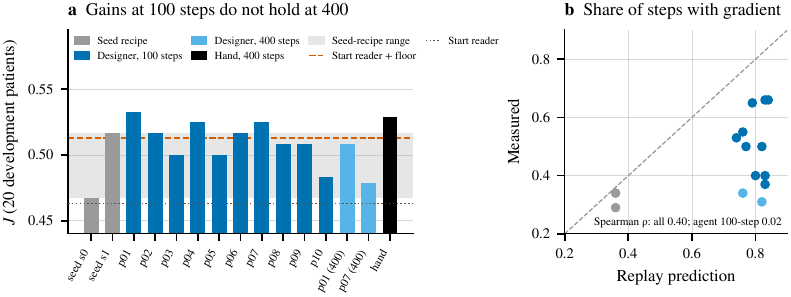}
\caption{
\textbf{Short-run recipe gains do not survive full training.}
\textbf{(a)} Development objective $J$ for seed, designer, and reference
recipes; the dashed threshold is the incumbent plus its measured noise floor.
\textbf{(b)} Replay-predicted versus measured fraction of training steps with
learning signal.
}
\label{fig:recipe_roundA}
\end{figure}

The search starts from the RL reader at $J=0.463$.
Two seed-recipe repeats establish a noise floor of 0.050, after which the
designer proposes ten recipes that are screened for 100 steps
(Table~\ref{tab:recipe_roundA}; Figure~\ref{fig:recipe_roundA}).
Three short trials appear to clear the acceptance threshold, but the two
strongest candidates fail to retain these gains when extended to 400 steps:
their improvements fall to $+0.046$ and $+0.017$, so neither is promoted.

The replay census is useful as a coarse screen but not as a predictor of final
performance. It overestimates the realized fraction of training steps carrying
learning signal, and its ranking of the designer's short-run proposals is
nearly uninformative (Spearman $\rho=0.02$). The hand-designed learnability
recipe reaches $J=0.529$ after full training and is included only as a
reference.

This round illustrates the separation between \emph{proposal} and
\emph{promotion} in ReVision3D: inexpensive screens and short trials can suggest
promising revisions, but only improvement on the fixed objective beyond the
measured noise floor is retained.

\subsection{Recipe Search After Perception Adaptation: No Further Gain}
\label{app:recipe-second}
% Second recipe search (values from the loop's trial registry).
\begin{table}[H]
\centering
\small
\setlength{\tabcolsep}{5pt}
\caption{\textbf{Second recipe search, from the vision-adapter reader} ($J=0.575$, never-detected share 0.30; 20 development patients,
100 steps per trial). Noise floor: range of the three seed-recipe runs, 0.042. No proposal improves on the start reader by more than the
floor, and the search is rejected. Never detected: share of lesions no slice read detects (Eq.~\eqref{eq:attribution}).}
\label{tab:recipe_second}
\begin{tabular}{llccc}
\toprule
Trial & Recipe by & $J$ & vs.\ start & Never detected \\
\midrule
seed $s_0$ & seed recipe & 0.567 & $-0.008$ & 0.27 \\
seed $s_1$ & seed recipe & 0.546 & $-0.029$ & 0.32 \\
seed $s_2$ & seed recipe & 0.525 & $-0.050$ & 0.23 \\
\midrule
p01 & designer & 0.583 & $+0.008$ & 0.37 \\
p02 & designer & 0.533 & $-0.042$ & 0.38 \\
p03 & designer & 0.554 & $-0.021$ & 0.27 \\
p04 & designer & \textbf{0.592} & $+0.017$ & 0.32 \\
\bottomrule
\end{tabular}
\end{table}

After accepting the perception revision, ReVision3D revisits the recipe level
from the stronger incumbent ($J=0.575$, Table~\ref{tab:recipe_second}).
Three seed-recipe repeats establish a noise floor of 0.042, and the designer
proposes four new recipes using the updated attribution.

None passes the promotion gate. The strongest proposal reaches $J=0.592$,
only $+0.017$ over the incumbent and well within measured seed variability.
Its never-detected share also remains within the range of the seed runs.
The round is therefore rejected.

This negative result is part of the recursive process: after perception changes,
attribution is recomputed and previously exhausted levels can be reconsidered.
Here, revisiting the recipe level produces no measurable gain, providing
evidence that the remaining loss is not addressed by further recipe changes.

\subsection{Learnability Census and Exhaustion of Recipe-Level Signal}

% Auto-generated by make_all_tables.py from RESULTS_ALL.tex -- edit the data, not this file.
\begin{table}[H]
\centering
\small
\setlength{\tabcolsep}{3.2pt}
\caption{
\textbf{Learnability census on training patients.}
Single-slice prompts are evaluated with eight samples at $T=0.9$.
\emph{Learnable} denotes prompts whose sampled rewards differ and therefore provide non-zero group-relative learning signal.
$^\ddagger$ denotes the training pool used for the second fresh-view reader.
}
\label{tab:census}
\begin{adjustbox}{max width=\linewidth}
\begin{tabular}{lr ccc ccc c}
\toprule
 & & \multicolumn{3}{c}{Lesion-bearing prompts} & \multicolumn{3}{c}{Lesion-free prompts} & \\
\cmidrule(lr){3-5}\cmidrule(lr){6-8}
Reader, prompt pool & Prompts & Always hit & Learnable & Never hit & Always empty & Sometimes FP & Always FP & Learnable \\
\midrule
RL reader, cached slices & 1{,}593 & 24\% & 37\% & 39\% & 84\% & 13\% & 3\% & \textbf{25\%} \\
RL reader, fresh views & 3{,}083 & 19\% & 31\% & 50\% & 84\% & 14\% & 1\% & \textbf{23\%} \\
Fresh-view reader, fresh views$^\ddagger$ & 2{,}797 & 27\% & 11\% & 62\% & 93\% & 4\% & 3\% & \textbf{7\%} \\
\bottomrule
\end{tabular}
\end{adjustbox}
\end{table}

\begin{figure}[H]
\centering
\includegraphics[width=\linewidth]{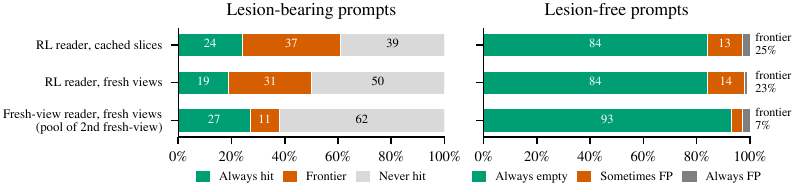}
\caption{
\textbf{Learnability census.}
Training prompts are grouped by the outcomes of eight sampled reads.
For lesion-bearing prompts, only the frontier group (i.e. where sampled rewards differ) provides group-relative learning signal.
The overall frontier share is shown on the right.
}
\label{fig:census}
\end{figure}

Table~\ref{tab:census} and Figure~\ref{fig:census} partition prompts by the outcomes of eight sampled reads. Lesion-bearing prompts are classified as always hit, frontier (sampled rewards differ), or never hit; only frontier prompts provide non-zero group-relative learning signal.

For the RL reader, 25\% of cached-view prompts and 23\% of fresh-view prompts lie on this frontier, while 39\% and 50\% of lesion-bearing prompts are never hit. After training on learnable fresh views, the frontier contracts to 7\% overall: more prompts become consistently solved, but 62\% of lesion-bearing prompts remain never hit. Lesion-free prompts also become more stable, with the always-empty share increasing from 84\% to 93\%.

The census explains why recipe-level optimization saturates. Training progressively consumes prompts that still exhibit reward variation, while never-hit prompts provide no group-relative signal. As the frontier contracts, additional recipe changes have diminishing leverage, motivating a move to the perception level.

\subsection{The Training Procedure Transfers Better Than the Reader}

% Auto-generated by make_all_tables.py from RESULTS_ALL.tex -- edit the data, not this file.
\begin{table}[H]
\centering
\small
\setlength{\tabcolsep}{4pt}
\caption{
\textbf{Transfer of the liver recipe to kidney.}
The same training procedure is applied unchanged to 30 held-out KiTS23 patients.
}
\label{tab:kidney_transfer}
\begin{adjustbox}{max width=\linewidth}
\begin{tabular}{ll cc cc}
\toprule
 & & \multicolumn{2}{c}{8-slice survey} & \multicolumn{2}{c}{+ lesion-guided slices} \\
\cmidrule(lr){3-4}\cmidrule(lr){5-6}
Reader & Training on kidney & R & FP & R & FP \\
\midrule
Liver reader, zero-shot & none on kidney & 0.16 {\scriptsize[0.03, 0.30]} & 0.37 & 0.23 & 0.53 \\
Kidney warm start & supervised & 0.48 {\scriptsize[0.31, 0.67]} & 0.40 & 0.58 & 0.30 \\
Kidney RL reader & mask-grounded reward & 0.58 {\scriptsize[0.41, 0.77]} & 0.33 & 0.61 & 0.30 \\
\bottomrule
\end{tabular}
\end{adjustbox}
\end{table}

\begin{figure}[H]
\centering
\includegraphics[width=0.55\linewidth]{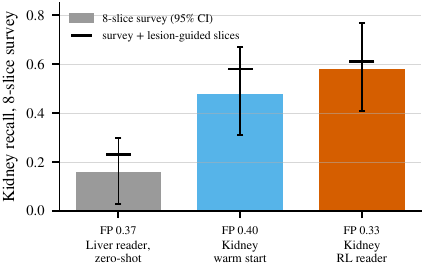}
\caption{
\textbf{Transfer from liver to kidney.}
Bars show recall on the 8-slice survey (95\% intervals); ticks show recall after adding lesion-guided slices.
False positives per patient are reported below each bar.
}
\label{fig:kidney_transfer}
\end{figure}

Table~\ref{tab:kidney_transfer} and Figure~\ref{fig:kidney_transfer} distinguish
transfer of the trained reader from transfer of the training procedure.
Applying the liver reader directly to kidney generalizes poorly, reaching
0.16 recall at 0.37 false positives per patient. In contrast, reusing the same
warm-start and mask-grounded RL procedure on kidney data reaches 0.48 recall
after supervised initialization and 0.58 at 0.33 false positives per patient
after RL.

Adding lesion-guided slices raises recall only from 0.58 to 0.61, suggesting
that limited view exposure explains little of the remaining error. Thus, in
this experiment, the training procedure transfers across organs better than the
resulting organ-specific reader: the same optimization recipe produces useful
readers for both liver and kidney, whereas the liver-trained reader itself does
not transfer directly.

\section{Successive Readers and Cascades}
\label{app:successive}

\subsection{Successive Recipe Revisions Reach a Similar Ceiling}

% Auto-generated by make_all_tables.py from RESULTS_ALL.tex -- edit the data, not this file.
\begin{table}[H]
\centering
\small
\setlength{\tabcolsep}{3.6pt}
\caption{
\textbf{Successive recipe-trained readers on 31 held-out liver patients.}
All results use dense reading with support $\geq2$.
$R$, FP, and $P$ denote lesion recall (95\% patient-bootstrap interval),
false positives per patient, and precision.
\emph{Zero-variance steps} is the fraction of training steps whose sampled
reward group is constant; \emph{Learnable} is the pre-training fraction of
candidate prompts whose eight sampled rewards differ.
Support is defined in Appendix~\ref{app:terms}.
}
\label{tab:successive}
\begin{adjustbox}{max width=\linewidth}
\begin{tabular}{ll cc c cc}
\toprule
Reader & Trained on & R & FP & P & Zero-variance steps & Learnable \\
\midrule
RL reader & mined failures & 0.45 {\scriptsize[0.34, 0.56]} & 0.52 & 0.76 & 19\% & -- \\
Longer-training control & un-mined & 0.41 {\scriptsize[0.28, 0.54]} & 0.39 & 0.79 & 88\% & -- \\
Failure-mined reader & failure-mined cases & 0.46 {\scriptsize[0.35, 0.55]} & 0.74 & 0.69 & 55\% & -- \\
Cached-view reader & learnable, cached views & 0.41 {\scriptsize[0.31, 0.51]} & \textbf{0.06} & \textbf{0.96} & 75\% & 25\% \\
Fresh-view reader & learnable, fresh views & \textbf{0.49} {\scriptsize[0.38, 0.59]} & 0.42 & 0.81 & 62\% & 23\% \\
Second fresh-view reader & learnable, fresh views & 0.45 {\scriptsize[0.33, 0.58]} & 0.32 & 0.83 & 53\% & 7\% \\
\bottomrule
\end{tabular}
\end{adjustbox}
\end{table}

\begin{figure}[H]
\centering
\includegraphics[width=\linewidth]{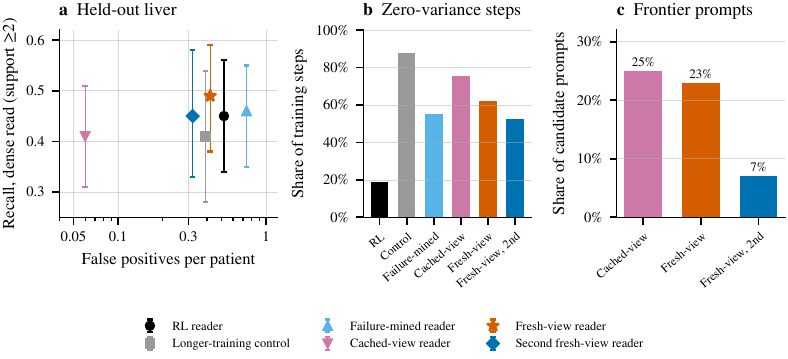}
\caption{
\textbf{Successive recipe-trained readers on held-out liver.}
\textbf{(a)} Dense-read recall and false positives at support $\geq2$
(95\% patient-bootstrap intervals).
\textbf{(b)} Fraction of training steps with zero reward variance.
\textbf{(c)} Fraction of candidate prompts on the learnable frontier before each
reader is trained.
}
\label{fig:operating_curves}
\end{figure}

Table~\ref{tab:successive} and Figure~\ref{fig:operating_curves} compare readers
produced by successive recipe revisions from the same RL reader. Despite
different curricula, dense-read recall remains within 0.41--0.49 with
overlapping confidence intervals, indicating a largely unchanged single-reader
detection ceiling.

The main differences are instead in precision. For example, the cached-view
reader reaches precision 0.96 at 0.06 false positives per patient, whereas the
failure-mined reader reaches 0.69 at 0.74. Meanwhile, the available learning
signal contracts: 53--88\% of training steps for post-RL readers have zero
reward variance, compared with 19\% for the RL reader, and the learnable
frontier shrinks from 25\% to 7\% of candidate prompts.

Thus, successive recipe revisions mainly move the reader along the
precision-recall trade-off while progressively exhausting the available
group-relative learning signal. This supports treating the recipe level as
exhausted.

\subsection{Successive Readers Provide Complementary Evidence}

% Auto-generated by make_all_tables.py from RESULTS_ALL.tex -- edit the data, not this file.
\begin{table}[H]
\centering
\small
\setlength{\tabcolsep}{3.4pt}
\caption{
\textbf{Cascades and paired reader comparisons on 31 held-out liver patients.}
Absolute columns report the first-named reader; $\Delta$R uses 95\% patient-bootstrap intervals and $\Delta$FP Wilcoxon signed-rank $p$-values.
Cascade proposals are retained when confirmed within 12\,mm in 3D.
}
\label{tab:cascade}
\begin{adjustbox}{max width=\linewidth}
\begin{tabular}{l cc cc cc}
\toprule
Comparison & R & FP & $\Delta$R & 95\% CI & $\Delta$FP & $p$ \\
\midrule
\multicolumn{7}{l}{\textit{(a) Cascade: the RL reader proposes, a later reader confirms (dense read)}} \\
RL reader alone & 0.45 {\scriptsize[0.34, 0.56]} & 0.52 & \multicolumn{2}{c}{reference} & \multicolumn{2}{c}{reference} \\
RL proposes, cached-view confirms & 0.56 {\scriptsize[0.44, 0.65]} & 0.61 & +0.11 & {\scriptsize[+0.03, +0.18]} & +0.10 & 0.821 \\
RL proposes, fresh-view confirms & 0.63 {\scriptsize[0.53, 0.72]} & 1.77 & +0.18 & {\scriptsize[+0.12, +0.24]} & +1.26 & $<$0.001 \\
RL proposes, second fresh-view confirms & 0.65 {\scriptsize[0.56, 0.73]} & 1.03 & +0.20 & {\scriptsize[+0.13, +0.26]} & +0.52 & 0.003 \\
\midrule
\multicolumn{7}{l}{\textit{(b) Paired reader comparisons, dense read at support $\geq$2}} \\
Fresh-view vs RL & 0.49 {\scriptsize[0.39, 0.59]} & 0.42 & +0.04 & {\scriptsize[$-$0.01, +0.08]} & $-$0.10 & 0.382 \\
Second fresh-view vs RL & 0.45 {\scriptsize[0.33, 0.57]} & 0.32 & +0.00 & {\scriptsize[$-$0.05, +0.06]} & $-$0.19 & 0.239 \\
Second fresh-view vs fresh-view & 0.45 {\scriptsize[0.33, 0.57]} & 0.32 & $-$0.04 & {\scriptsize[$-$0.08, +0.01]} & $-$0.10 & 0.382 \\
Cached-view vs longer-training control & 0.41 {\scriptsize[0.30, 0.50]} & 0.06 & +0.00 & {\scriptsize[$-$0.06, +0.06]} & $-$0.32 & 0.085 \\
Fresh-view vs longer-training control & 0.49 {\scriptsize[0.38, 0.59]} & 0.42 & +0.08 & {\scriptsize[+0.01, +0.14]} & +0.03 & 0.746 \\
Fresh-view vs cached-view & 0.49 {\scriptsize[0.39, 0.59]} & 0.42 & +0.08 & {\scriptsize[+0.04, +0.12]} & +0.35 & 0.137 \\
\midrule
\multicolumn{7}{l}{\textit{(c) Paired reader comparisons, 8-slice survey}} \\
Fresh-view vs RL & 0.31 {\scriptsize[0.19, 0.45]} & 0.68 & +0.04 & {\scriptsize[$-$0.01, +0.09]} & $-$0.29 & 0.380 \\
Second fresh-view vs RL & 0.30 {\scriptsize[0.19, 0.41]} & 0.29 & +0.03 & {\scriptsize[$-$0.02, +0.08]} & $-$0.68 & 0.073 \\
\bottomrule
\end{tabular}
\end{adjustbox}
\end{table}

\begin{figure}[H]
\centering
\includegraphics[width=\linewidth]{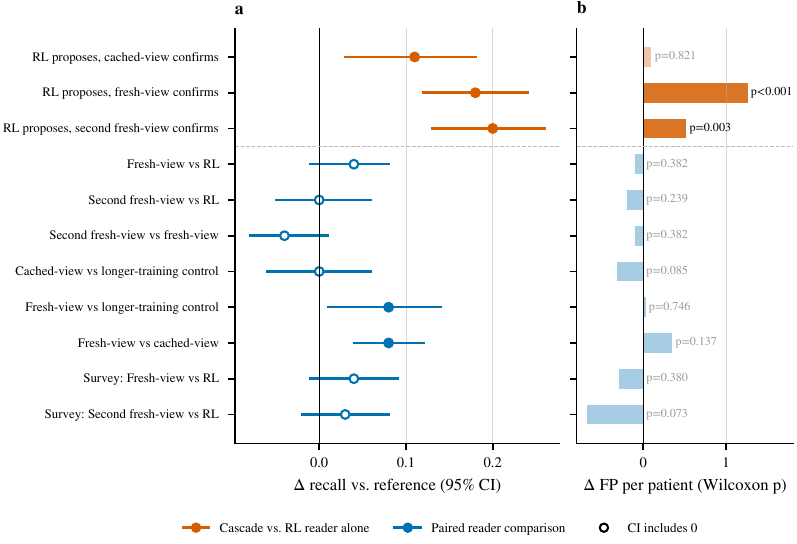}
\caption{
\textbf{Complementarity between successive readers on held-out liver.}
\textbf{(a)} Paired change in recall with 95\% patient-bootstrap intervals;
hollow markers indicate intervals containing zero.
\textbf{(b)} Paired change in false positives per patient; saturated bars denote
Wilcoxon $p<0.05$.
}
\label{fig:cascade_forest}
\end{figure}

Although successive readers reach similar recall individually, their errors are
not identical. Table~\ref{tab:cascade} and Figure~\ref{fig:cascade_forest}
measure this complementarity using proposal-confirmation cascades and paired
reader comparisons on the same patients.

Using the cached-view reader to confirm the RL reader increases recall by
0.11 [0.03, 0.18] without a significant increase in false positives
($p=0.82$). Fresh-view confirmation yields larger gains of 0.18--0.20, but at
the cost of 0.52--1.26 additional false positives per patient. Among individual
readers, only the fresh-view reader significantly improves over the
equal-compute control (+0.08 [0.01, 0.14]) and the cached-view reader
(+0.08 [0.04, 0.12]); the remaining pairwise recall differences include zero.

Successive readers are therefore more useful as \emph{complementary evidence}
than as replacements for one another. This motivates the inference-level
revision in ReVision3D, which combines findings across readers rather than
selecting a single final checkpoint.

\section{Designer Models and Ablations}
\label{app:designers}

\subsection{The Improvement Loop Is Robust Across Tested Designers}
% Auto-generated by make_all_tables.py from RESULTS_ALL.tex -- edit the data, not this file.
\begin{table}[H]
\centering
\small
\setlength{\tabcolsep}{3.2pt}
\caption{
\textbf{Inference-level improvement is consistent across tested designers.}
Four language-model designers are evaluated from the same three incumbents.
$J$ is the accepted development objective and $\Delta J$ its gain over the
incumbent. \emph{Rev.} denotes the accepted revision index among up to eight
proposals; acceptance requires $P(\Delta J>0)\geq0.9$.
}
\label{tab:designers}
\begin{adjustbox}{max width=\linewidth}
\begin{tabular}{l ccc ccc ccc}
\toprule
 & \multicolumn{3}{c}{Liver, liver-study readers} & \multicolumn{3}{c}{Liver, multi-organ readers} & \multicolumn{3}{c}{Kidney, multi-organ readers} \\
 & \multicolumn{3}{c}{\scriptsize incumbent $J=0.575$} & \multicolumn{3}{c}{\scriptsize incumbent $J=0.679$} & \multicolumn{3}{c}{\scriptsize incumbent $J=0.801$} \\
\cmidrule(lr){2-4}\cmidrule(lr){5-7}\cmidrule(lr){8-10}
Designer & $J$ & $\Delta J$ & Rev. & $J$ & $\Delta J$ & Rev. & $J$ & $\Delta J$ & Rev. \\
\midrule
Claude Opus 5 & 0.608 & +0.033 & 1 & \textbf{0.750} & +0.071 & 2 & 0.881 & +0.080 & 1 \\
GPT-6 Astra & \textbf{0.613} & +0.038 & 1 & 0.746 & +0.067 & 1 & 0.881 & +0.080 & 1 \\
Gemini 3.1 Pro & \textbf{0.613} & +0.038 & 1 & 0.746 & +0.067 & 1 & \textbf{0.892} & +0.091 & 1, 2 \\
Gemini 3.8 Flash & \textbf{0.613} & +0.038 & 1 & \textbf{0.750} & +0.071 & 1 & 0.886 & +0.085 & 1 \\
\midrule
Spread across designers & 0.005 &  &  & 0.004 &  &  & 0.011 &  &  \\
\bottomrule
\end{tabular}
\end{adjustbox}
\end{table}

\label{app:designers-ablations}

To test sensitivity to the language-model designer, Table~\ref{tab:designers}
repeats the inference-level search with four designers from three model families
and three different incumbents ($J=0.575$, $0.679$, and $0.801$).

All twelve runs produce an accepted program, eleven on the first revision.
Within each incumbent setting, accepted objectives differ across designers by at
most 0.005, 0.004, and 0.011, respectively. Gemini 3.8 Flash also performs
comparably as a designer despite being substantially weaker as a medical image
reader under the same evaluation protocol.

Across the tested designers, inference-level improvement is therefore stable to
the model proposing revisions. The designer generates candidate programs, while
the fixed external objective determines which candidates are retained.

\subsection{Ablations Localize the Largest Gain to Perception}
% Auto-generated by make_all_tables.py from RESULTS_ALL.tex -- edit the data, not this file.
\begin{table}[H]
\centering
\small
\setlength{\tabcolsep}{4pt}
\caption{
\textbf{Component-wise ablations on the same 31 held-out liver patients.}
Each block varies one intervention while holding the others fixed:
(a) training signal, (b) reading protocol, (c) training recipe, and
(d) perception configuration.
Recall is stratified by lesion volume; FP denotes false positives per patient.
Block (a) uses the 8-slice survey, while blocks (b)--(d) use the dense protocol
at $\leq0.5$ FP per patient or support $\geq2$ as indicated.
Bold denotes the best result within each block.
}
\label{tab:ablation_main}
\begin{adjustbox}{max width=\linewidth}
\begin{tabular}{l cccc cc}
\toprule
 & \multicolumn{4}{c}{Lesion volume (mL)} & & \\
\cmidrule(lr){2-5}
Setting (liver, 31 test patients) & 1--5 & 5--20 & 20--100 & $>$100 & All & FP \\
{\scriptsize\textit{Number of lesions}} & {\scriptsize52} & {\scriptsize31} & {\scriptsize17} & {\scriptsize11} & {\scriptsize111} & \\
\midrule
\multicolumn{7}{l}{\textit{(a) Training signal: our 4B reader after each training stage, 8-slice survey}} \\
Untrained Qwen3-VL-4B (zero-shot) & 0.02 & 0.03 & 0.06 & 0.18 & 0.05 & 1.32 \\
+ supervised warm start on mask-derived reports & 0.08 & 0.23 & 0.47 & \textbf{0.82} & 0.25 & 0.42 \\
+ RL with the mask-grounded reward (the RL reader) & 0.08 & \textbf{0.26} & 0.53 & \textbf{0.82} & 0.27 & 0.97 \\
\quad control: 10 supervised epochs instead of RL & \textbf{0.12} & 0.19 & \textbf{0.65} & \textbf{0.82} & \textbf{0.29} & 1.71 \\
\midrule
\multicolumn{7}{l}{\textit{(b) Reading protocol, RL reader}} \\
8 uniform slices & 0.08 & 0.26 & 0.53 & 0.82 & 0.27 & 0.97 \\
dense read (every slice) & 0.29 & 0.39 & 0.76 & \textbf{0.91} & 0.45 & 0.52 \\
+ a second reader confirms & \textbf{0.37} & \textbf{0.58} & \textbf{0.88} & \textbf{0.91} & \textbf{0.56} & 0.61 \\
\midrule
\multicolumn{7}{l}{\textit{(c) Training recipe: readers trained from the RL reader, dense read}} \\
RL reader (reference) & 0.29 & 0.39 & 0.76 & \textbf{0.91} & 0.45 & 0.52 \\
\quad 400 more steps, no selection & 0.27 & 0.29 & 0.71 & \textbf{0.91} & 0.41 & 0.39 \\
\quad trained on its failures & \textbf{0.31} & 0.39 & 0.76 & \textbf{0.91} & 0.46 & 0.74 \\
\quad learnable prompts, cached views & 0.23 & 0.45 & 0.59 & 0.82 & 0.41 & 0.06 \\
\quad learnable prompts, fresh views & \textbf{0.31} & 0.45 & \textbf{0.82} & \textbf{0.91} & \textbf{0.49} & 0.42 \\
\quad learnable prompts, fresh views, repeated & 0.25 & \textbf{0.48} & 0.76 & 0.82 & 0.45 & 0.32 \\
\midrule
\multicolumn{7}{l}{\textit{(d) Perception configuration: multi-organ reader, dense read}} \\
Frozen vision encoder & 0.06 & 0.13 & 0.41 & 0.64 & 0.19 & 0.00 \\
Adapters, 4 blocks (human-chosen) & 0.12 & 0.19 & 0.53 & 0.64 & 0.25 & 0.06 \\
Adapters, 24 blocks (designer): \textit{ReVision3D} & \textbf{0.69} & \textbf{0.84} & \textbf{1.00} & \textbf{0.82} & \textbf{0.79} & 0.35 \\
\bottomrule
\end{tabular}
\end{adjustbox}
\end{table}

Table~\ref{tab:ablation_main} changes one component at a time on the 31 held-out
liver patients and reports recall by lesion volume. The pattern separates the
effects of training signal, reading protocol, training recipe, and perception.

A supervised warm start raises overall survey recall from 0.05 to 0.25, mainly
for lesions above 20\,mL, while RL and a 10-epoch supervised control leave
1--5\,mL recall at only 0.08--0.12. Dense reading raises small-lesion recall
from 0.08 to 0.29, and confirmation by a second reader reaches 0.37. Recipe
variants remain within 0.23--0.31 on the same bin. In contrast, changing the
perception configuration raises 1--5\,mL recall from 0.06 with the frozen encoder
and 0.12 with hand-selected adapters to 0.69, while overall recall rises from
0.19 and 0.25 to 0.79.

The ablations are consistent with the attribution analysis: acquisition and
inference recover or reconcile evidence already within reach, recipe changes
primarily alter how that evidence is used, and perception adaptation produces
the largest increase in the set of lesions detected by the reader.

\section{Extended Limitations and Future Directions}
\label{app:limitations}

\paragraph{Dependence on an external verifier.}
ReVision3D can optimize only outputs that can be checked independently of the
reader. In this work, reference lesion masks provide such a verifier for spatial
localization, but not for characterization, malignancy, etiology, or other
clinical judgments. The framework therefore improves a verifiable component of
medical perception rather than clinical reasoning as a whole. A natural
extension is to broaden the verifier beyond spatial masks by incorporating
additional externally grounded signals, such as longitudinal follow-up,
pathology, structured measurements, or cross-modality agreement, so that more
dimensions of clinical interpretation can enter the improvement loop.

\paragraph{Bounded intervention space.}
The designer searches only four predefined levels: acquisition, perception
configuration, training recipe, and inference. This makes revisions auditable
and comparable under a fixed objective, but also limits what the system can
discover. The current loop cannot introduce a new backbone, imaging modality,
rendering operator, or downstream reasoning strategy unless that capability is
explicitly exposed as an action. Future systems could progressively expand
their own intervention space, for example by proposing new visual modules,
rendering strategies, or tool interfaces and admitting them only after the
same external promotion test.

\paragraph{Replay is exact for geometry, not perception.}
Volumetric replay can render unvisited views exactly and determine their
geometric lesion exposure from reference masks, but it cannot know how the
reader will respond to an unseen view without running the model. Acquisition
and inference can therefore be screened cheaply, whereas perception revisions
still require training and evaluation. A promising direction is to learn
lightweight surrogates that predict which perception changes are likely to
reduce the never-detected share, using previous trials as experience while
retaining the fixed evaluator for final promotion.

\paragraph{Attribution guides intervention but is not causal.}
The missed-lesion decomposition is exact, but the split between training-level
and perception-level loss relies on finite sampled reads and the learnability
census. It is therefore a routing heuristic rather than a causal guarantee.
Future work could make routing itself adaptive, for example by explicitly
estimating the expected improvement of each intervention from prior revisions
or by allocating a small exploration budget across multiple levels before
committing expensive training compute.

\paragraph{A fixed objective defines the boundary of improvement.}
Keeping the verifier and system objective fixed is important for preventing the
designer from redefining success, but it also constrains what counts as
progress. Improvements in lesion localization may not always translate to
better downstream clinical decisions. An important next step is to study
hierarchical objectives in which spatially verifiable perception remains the
grounded inner loop, while higher-level clinical utility is evaluated
separately under independent evidence and human oversight.

\end{document}